%% file: SplatFlow-Submission-For-IJCV.tex
\documentclass[twocolumn]{svjour3}          
\input{body/config}

\begin{document}

\input{body/head}
\input{body/abstract}

\section{Introduction}

Optical flow estimation (OFE) aims to estimate the pixel-wise 2D motion between consecutive video frames. As a longstanding, fundamental yet challenging problem in low-level vision \cite{horn1981determining, perception}, OFE has been extensively studied for several decades due to its numerous applications, including video object segmentation \cite{cheng2017segflow}, video object tracking \cite{aslani2013optical}, VSLAM \cite{choi2014multi}, action recognition \cite{sevilla2018integration}, 3D reconstruction \cite{li2019learning}, video inpainting \cite{xu2019deep, gao2020flow}, video superresolution \cite{sajjadi2018frame}, depth estimation \cite{ranftl2016dense}, and scene flow estimation \cite{yang2020upgrading}. 
Before the advent of deep learning, variational approaches had dominated the OFE field since the work by Horn and Schunck \cite{horn1981determining}.

Recently, akin to many vision problems, the boom of deep learning has injected new vitality into the OFE field, causing the focus to transfer from traditional methods \cite{brox2009large, horn1981determining, xu2011motion} to deep learning techniques \cite{Flownet_and_Chairs, Flownet2, PWCNet, RAFT}. Indeed, deep learning-based methods like PWC-Net \cite{PWCNet} and RAFT \cite{RAFT} have achieved higher accuracy than the best classical methods \cite{revaud2015epicflow, weinzaepfel2013deepflow, bailer2015flow} while being incredibly faster. However, most OFE methods face challenging problems, such as a high computational cost and limited deployment on edge devices. These challenges, as discussed below, must be overcome before OFE methods can be applied to real-world scenarios:

\begin{itemize}
\renewcommand{\labelitemi}{$\bullet$}
  \item \textbf{High computational cost.} 
  Most recent deep learning-based methods \cite{RAFT, GMA, zhang2021separable} perform OFE based on a correlation volume that stores the pairwise pixel similarities between two frames. However, the computation grows quadratically with the number of pixels in a frame \cite{jiang2021learning}. In addition, most such methods are constrained to be deployed in resource-limited edge devices.
  \item \textbf{High cost of collecting large-scale labeled data.} 
  Collecting large-scale, densely labeled, realistic data for OFE is also prohibitively costly. Most current methods use synthetic data \cite{Things, Sintel} for training, which decreases accuracy when dealing with realistic data due to the domain gap \cite{liu2019selflow}.
  Therefore, unsupervised \cite{wang2018occlusion, jonschkowski2020matters} or self-supervised \cite{liu2019selflow} methods have been explored to address this challenge.
  \item \textbf{Occlusion.} 
  Proper occlusion handling is critical for accurate OFE, as occlusion is inevitable. Matching costs are inaccessible in occluded regions, which significantly impacts OFE accuracy. 
\end{itemize}

Occlusion significantly limits the application of deep learning-based methods for OFE. 
Accurate OFE in occluded regions is highly beneficial for high-level visual tasks, as it allows for a better understanding of the spatial relationships between objects in the scene and helps to realize the complete motion tracking of objects. This study aims to address the occlusion problem in a careful and considerate way.

Existing popular deep learning-based OFE methods rely on two consecutive frames, which can be quite limiting when handling occlusions. Under the two-frame settings, the networks directly match or indirectly regress each pixel's optical flow based on the significant feature similarity between pixel pairs.
Unfortunately, this approach is not feasible in occluded regions. The occluded regions will be occluded by other regions in the next frame or even move out of the image boundary. In this case, the feature similarity values will generally be low for occluded regions due to no good correspondences, leading to failed estimations.
Although Jiang \etal~\cite{GMA} proposed that the flow in non-occluded regions in the current frame can be aggregated by feature similarity, which can be used as a reference for estimating the optical flow in occluded regions, this method can only solve a tiny portion of the occlusions. In most cases, occluded regions cannot find a reference with high feature similarity in the current frame.

As discussed earlier, handling occlusions is very challenging under the two-frame settings. One straightforward way is to consider multiple frames by exploring temporal smoothness. While this approach offers richer information for exploring various assumptions and can potentially mitigate the occlusion issue, it introduces multiple difficulties, such as potential harm during dramatic changes in the flow field \cite{volz2011modeling}, error accumulation and diffusion from the previous frame's estimation, and higher computational costs \cite{irani1999multi, wulff2017optical}. Unfortunately, the multi-frame optical flow estimation (MOFE) problem remains underexplored, with only a handful of studies addressing the issues \cite{ren2019fusion, neoral2018continual, liu2019selflow} with the following limitations and the following limitations to overcome.
  
\begin{itemize}
\renewcommand{\labelitemi}{$\bullet$}
  \item \textbf{Few specialized multi-frame methods are available for the more recent single-resolution iterative two-frame backbones.}
  There are two types of backbones for OFE; pyramidal backbones, such as PWC-Net \cite{PWCNet} and LiteFlowNet \cite{hui2018liteflownet}, and single-resolution iterative backbones, such as RAFT \cite{RAFT} and GMA \cite{GMA}. Existing deep learning-based multi-frame methods for OFE tend to fall into two categories, those specially designed for pyramidal backbones \cite{liu2019selflow, neoral2018continual} and those agnostic to backbones \cite{ren2019fusion, maurer2018proflow}. The former category cannot adapt to the recent single-resolution iterative backbones, such as RAFT \cite{RAFT} and GMA \cite{GMA}. 
  The latter category is agnostic to backbone architectures and thus cannot consider the advantages of different backbone architectures. For instance, FusionFlow \cite{ren2019fusion} cannot fuse the previous frame's motion estimates into each layer of PWC-Net \cite{PWCNet} or each iteration of RAFT \cite{RAFT}. 
  Therefore, they can only achieve sub-optimal estimation performance or sometimes can not outperform two-frame methods. As far as we know, the WarmStart initialization proposed in RAFT \cite{RAFT} is the only method specially designed for the single-resolution iterative backbones, but it still struggles with occlusions.
  \item \textbf{Limitations of the commonly used feature alignment methods.} 
  The previous frame has a different coordinate system than the current frame. Most existing methods obtain the aligned features from the previous frame, such as correlation volume and optical flow, which will be embedded into the two-frame backbones by executing the inference process of backward flow from the current frame to the previous frame or performing a non-differentiable forward warping transformation. However, as a most direct way, computing the backward flow will increase the computational cost at least twice, seriously affecting the algorithm's real-time performance. 
  Moreover, using non-differentiable forward warping transformation will inhibit the network from propagating gradients through this step, hindering the effective learning of multi-frame knowledge during training.
\end{itemize}

This paper proposes SplatFlow, a straightforward yet efficient MOFE framework for the occlusion problem. 
The novelty of our method stands on careful consideration of the unique ``single-resolution'' characteristic of the current single-resolution iterative two-frame backbones, the downsides of high cost in backward flow calculation, and the limitations of non-differentiability in the commonly used forward warping transformation.
The proposed SplatFlow introduces the differentiable splatting transformation to align the previous frame's motion feature and designs a Final-to-All embedding method for inputting the aligned motion feature into the current frame's estimation, thereby remodeling the existing two-frame backbones.

Specifically, our proposed SplatFlow framework propagates the motion estimates from a previous frame to the current frame. This framework encompasses the encoding, alignment, and embedding of the motion feature. 
First, we use the motion feature from RAFT \cite{RAFT}, which is the convolutional coding feature of the flow and the correlation feature, to represent the previous frame's motion estimates. 
Second, we propose a Splatting-based motion feature alignment method to replace backward flow calculation or non-differentiable forward warping transformation. Here we introduce the splatting transformation widely used in video frame interpolation into the MOFE application. The previous frame's motion feature is injected with bi-linear weights into the adjacent four pixels of the target sub-pixel position determined by the previous frame's optical flow, resulting in a differentiable and forward motion feature alignment at the sub-pixel level.
Finally, we specially design a Final-to-All motion feature embedding method for the single-resolution iterative two-frame backbones. Specifically, the network selects the previous frame's final iteration's aligned motion feature (\textbf{Final}) and further inputs it into every iteration (\textbf{All}) of the current frame's estimation process.

The SplatFlow framework is highly adapted to the single-resolution iterative two-frame backbone networks, such as RAFT \cite{RAFT} and GMA \cite{GMA}. Compared with the original two-frame backbone, our method has significantly higher estimation accuracy, especially in occluded regions, while maintaining a comparable inference speed. At the time of submission, this method achieved state-of-the-art results on the challenging Sintel benchmark \cite{Sintel}, with surprisingly significant 19.4\% (clean pass) and 16.2\% (final pass) error reductions compared to the previous best result submitted. Our method also outperformed all published pure optical flow methods on the KITTI2015 benchmark \cite{KITTI}.

This study offers the following contributions:

\begin{itemize}
\renewcommand{\labelitemi}{$\bullet$}
    \item We propose a novel MOFE framework SplatFlow designed explicitly for the single-resolution iterative two-frame backbones. Compared with the original backbone, SplatFlow has significantly higher estimation accuracy, especially in occluded regions, while maintaining a high inference speed.
    \item In our framework, we propose a Splatting-based motion feature alignment method to replace the commonly used backward flow calculation or non-differentiable forward warping transformation, thus achieving a differentiable and forward motion feature alignment at the sub-pixel level.
    \item Additionally, we specially design a Final-to-All motion feature embedding method to fully use the backbone's unique ``single-resolution'' characteristic.
    \item At the time of submission, our SplatFlow achieved state-of-the-art results on both the Sintel \cite{Sintel} and KITTI2015 \cite{KITTI} benchmarks, especially with surprisingly significant 19.4\% (clean pass) and 16.2\% (final pass) error reductions compared to the previous best result submitted on the Sintel \cite{Sintel} benchmark.
\end{itemize}

\section{Related Work}

\subsection{Two-frame Deep Learning Methods}

Thanks to the advance of deep neural networks (DNNs) in various computer vision tasks, the OFE methods based on deep learning gradually surpass the traditional methods both in accuracy and speed.
FlowNet \cite{Flownet_and_Chairs} is the first method using DNNs to estimate optical flow.
FlowNet2.0 \cite{Flownet2} performs on par with the state-of-the-art methods by stacking a series of warping operations.
After that, a series of methods, such as PWC-Net \cite{PWCNet} and LiteFlowNet \cite{hui2018liteflownet}, adopt a pyramid structure to refine the estimated optical flow from coarse to fine.
However, this coarse-to-fine cascade has several limitations, such as the difficulty of recovering from errors at coarse resolutions and the tendency to miss small fast-moving objects, as described and addressed in RAFT \cite{RAFT}. RAFT \cite{RAFT} adopts an iterative method to continuously refine the optical flow results at the same high resolution and has achieved significant performance improvements.
Since then, many works \cite{GMA, jiang2021learning, xu2021high} have optimized this single-resolution iterative two-frame backbone to reduce computational costs or improve accuracy.
Among them, GMA \cite{GMA} is dedicated to solving occlusions under the two-frame settings, using the attention mechanism to capture motion clues from similar areas in the spatial domain corresponding to the current frame. However, this approach can only solve special cases in a tiny portion of occlusions.

\subsection{Multi-frame Deep Learning Methods}

Some studies enable the network to extract and fuse features from the previous frame under the multi-frame settings to deal with occlusions more comprehensively.

FusionFlow \cite{ren2019fusion} non-iteratively optimizes the two-frame method's full-resolution optical flow results using the previous frame's image and flow and does not depend on specific backbone network types. This method has good generality and is suitable for all kinds of two-frame backbones. However, it only performs post-processing, \ie, fusion operation, for the two-frame estimation results. As a result, it can only achieve sub-optimal performance compared to the specialized multi-frame methods. Proflow \cite{maurer2018proflow}, which is closely related to FusionFlow \cite{ren2019fusion}, has also been proposed, but its algorithm cannot outperform two-frame methods on many datasets.

For a coarse-to-fine network structure, SelFlow \cite{liu2019selflow} uses the forward correlation volume of the previous frame and the backward low-resolution optical flow to refine each layer's OFE result in a self-supervised manner. Similarly, ContinualFlow \cite{neoral2018continual} also embeds the previous frame's features into restoring a higher-resolution optical flow. However, this approach differs from other methods, using forward and backward warped optical flow instead of the forward correlation volume. 
According to the ``pyramidal'' characteristic of the pyramid backbone networks, the above methods cleverly embed the previous frame's features into refining the resolution from coarse to fine, remarkably improving the network's utilization effect of the multi-frame settings. 
However, the aforementioned multi-frame methods are not directly applicable to recent single-resolution iterative two-frame networks such as RAFT \cite{RAFT} and GMA \cite{GMA}. 

Based on our current knowledge, the WarmStart initialization proposed in RAFT \cite{RAFT} seems to be the only method specially designed for single-resolution iterative two-frame backbone networks. It forward projects the previous frame's optical flow to the current frame to replace zero as the initial value of the optical flow iterations. However, this early injection of the previous frame's optical flow is only a one-time event and will be forgotten after many GRU iterations. As a result, it still cannot effectively handle occlusions.

\subsection{Warping Transformations}

\input{fig/overview}

Given two adjacent image frames $I_1$ and $I_2$, the backward warping transformation refers to mapping $I_2$ using the optical flow from $I_1$ to $I_2$ through an image sampling operation \cite{jason2016back} to reconstruct $I_1$. The commonly used sampling mode is bi-linear sampling. The backward warping transformation is differentiable and is widely used in unsupervised optical flow estimation \cite{meister2018unflow, wang2018occlusion}, supervised optical flow estimation \cite{PWCNet, hui2018liteflownet}, unsupervised depth estimation \cite{godard2017unsupervised, mahjourian2018unsupervised}, novel view synthesis \cite{cun2018depth, liu2018geometry}, video frame interpolation \cite{bao2019depth, jiang2018super}, and video enhancement \cite{tao2017detail, caballero2017real}.

When a backward warping transformation is applied to multi-frame optical flow methods \cite{neoral2018continual, ren2019fusion}, the backward flow from $I_2$ to $I_1$ is generally used to map the image, flow, or other features corresponding to $I_1$ to align with $I_2$'s coordinate system. This approach allows the backward warped features to be input into the network for estimating the optical flow of the current frame $I_2$.

Forward warping transformation is mainly used in OFE \cite{wang2018occlusion, neoral2018continual} and video frame interpolation \cite{niklaus2018context, bao2019depth, niklaus2020softmax}. In forward warping transformation, the feature's mapping direction is consistent with the optical flow's direction, which differs from backward warping transformation. Specifically, when mapping $I_1$ to $I_2$, the optical flow from $I_1$ to $I_2$ provides targets' positions rather than the flow from $I_2$ to $I_1$.

The consistency of directions brings challenges to forward warping transformation, such as the conflict of mapping multiple sources to the same target and non-differentiability problems. ContinualFlow \cite{neoral2018continual} retains the maximum flow value at the same mapped target during mapping. However, since this warping is not differentiable, the network cannot propagate gradients through this step during training. Therefore, this warping proved useless in the experiments. The same problem occurs with CtxSyn \cite{niklaus2018context}, which uses the equivalent of z-buffering. In unsupervised optical flow estimation, Wang \etal \cite{wang2018occlusion} propose a differentiable method that takes the sum of contributions from all source pixels to the same target as the mapped value, which is known as splatting and is widely used in video frame interpolation \cite{niklaus2020softmax, hu2022many, siyao2021deep}.

To the best of our knowledge, we are the first to introduce the differentiable splatting transformation into MOFE.

\subsection{Point Tracking}

OFE aims to estimate the corresponding locations of interest points in the adjacent next frame, focusing more on the local instantaneous motion of points in the video sequence. In contrast, the point tracking task aims to estimate the locations and occlusion statuses of interest points in each video sequence frame, focusing more on the global motion trajectory at the point level. Similarly, the object tracking task \cite{ye2022joint, bhat2020know} focuses on a global motion from the object level.  
PIPs \cite{pips} combines cost volumes and iterative inference with a deep temporal network, which jointly reasons about the location and appearance of visual entities across multiple timesteps, achieving excellent tracking results. 
Further research \cite{tapnet, mft, tapir} has been done on the point tracking task. 
It is worth noting that our method's ``multi-frame'' is reflected in utilizing multi-frame information to estimate better the local instantaneous motion rather than the global motion trajectories that point tracking methods, such as PIPs \cite{pips}, focus on.

\section{Proposed Method}

In this section, the holistic architecture of our method is described in Section \ref{subsection:overall}. Then, the key contributions of our method, including the Splatting-based motion feature alignment method and the Final-to-All motion feature embedding method, are depicted in Sections \ref{subsection:splat} and \ref{subsection:o2m}, respectively. Some details of the model during training and inference are provided in Section \ref{subsection:training}.

\subsection{Overall Architecture of SplatFlow} \label{subsection:overall}

Our multi-frame method is designed explicitly for single-resolution iterative two-frame backbone networks such as RAFT \cite{RAFT} and GMA \cite{GMA}. We take RAFT \cite{RAFT} as an example to illustrate our method's overall architecture, as shown in Fig. \ref{fig:overview}.

The thick orange and purple arrows in Fig. \ref{fig:overview} represent the OFE processes $P_{t-1\rightarrow t}$ (from the frame $t-1$ to $t$) and $P_{t\rightarrow t+1}$ (from the frame $t$ to $t+1$) of the original RAFT \cite{RAFT}, respectively. The two processes have the same network structure and data flow except for different inputs. The $P_{t\rightarrow t+1}$ process takes two adjacent images, $I_{t}$ and $I_{t+1}$, as inputs and extracts the image feature $F^{i}$ and the context feature $F^{c}$ of each image with a Feature Encoder and Context Encoder networks that have identical structures but independent parameters. The Feature/Context Encoder network comprises 6 residual blocks and 3 downsampling layers, producing a dense feature map at 1/8 the input image resolution. The 4-layer correlation pyramid $Y_{t}$ between $F^{i}_{t}$ and $F^{i}_{t+1}$ is calculated by taking the dot product between all pairs of feature vectors and pooling the last two dimensions with kernel sizes 1, 2, 4, and 8 and equivalent stride.
Then, $Y_{t}$ and $F^{c}_{t}$ are used as inputs to update the low-resolution optical flow estimates $O_{t}$ with an initial value of 0 through an iterative method, which is shown as a single-resolution iteration module (SIM) in the purple part of Fig. \ref{fig:overview}, and the full-resolution result $O^{f}_{t}$ is recovered by using convex upsampling \cite{RAFT}. 
During the $n$th iteration of the GRU Predictor network in SIM, the correlation feature $C_{t,n}$ is generated by dynamically looking up $Y_{t}$ with $O_{t,n-1}$ and is encoded with $O_{t,n-1}$ to obtain the motion feature.
Then, a separable convolutional GRU Cell is used to input $x_n$, 
which is the concatenation of $F^{c}_{t}$ and the motion feature, to update the hidden state feature $h_n$:
\begin{align}
    & z_n = \sigma(\operatorname{Conv}([h_{n-1}, x_{n}], W_z)), \label{eq:gru_01} \\
    & r_n = \sigma(\operatorname{Conv}([h_{n-1}, x_{n}], W_r)), \label{eq:gru_02} \\
    & \tilde{h}_n = \tanh(\operatorname{Conv}([r_n \odot h_{n-1}, x_{n}], W_h)), \label{eq:gru_03} \\
    & h_n = (1 - z_n) \odot h_{n-1} + z_n \odot \tilde{h}_n. \label{eq:gru_4}
\end{align}
After that, $h_n$ is passed through two convolutional layers to predict a low-resolution optical flow residual $\Delta O_{t,n}$ that can be used to update $O_{t,n-1}$ to $O_{t,n}$.

Our SplatFlow multi-frame optical flow framework propagates the motion estimates from the $P_{t-1\rightarrow t}$ process to the $P_{t\rightarrow t+1}$ process. 
The SplatFlow framework first extracts the motion feature $M_{t-1,n}$ after each iteration of the $P_{t-1\rightarrow t}$ process and takes a Splatting-based alignment method to obtain the motion feature aligned with the frame $t$'s coordinate system, \ie, $A_{t-1,n}$. Then the framework uses a Final-to-All embedding method to input the aligned motion feature into the $P_{t\rightarrow t+1}$ estimation process.

The green part in Fig. \ref{fig:overview} shows our framework's Motion Feature Encoder network. We introduce the motion feature from the two-frame method RAFT \cite{RAFT}. Specifically, the network jointly encodes the $C_{t-1,n}$ and $O_{t-1,n-1}$ generated in each iteration of the $P_{t-1\rightarrow t}$ process and obtains $M_{t-1,n}$. Figure \ref{fig:mf} depicts the detailed $M_{t-1,n}$ generation process.

\input{fig/mf}

\input{fig/splatting}

We use the Splatting-based Aggregator network to implement our proposed Splatting-based motion feature alignment method, as shown in the blue part in Fig. \ref{fig:overview}.
After extracting the $M_{t-1,n}$ of each iteration, we use $O_{t-1,n}$ to unidirectionally map it to the coordinate system of the frame $t$ to obtain $A_{t-1,n}$. The motion feature's differentiable and sub-pixel level filling can be realized based on the splatting transformation.

In Fig. \ref{fig:overview}, the red part shows the Final-to-All Embedder network in our framework. 
The aligned motion feature $A_{t-1,n}$ of the final iteration is input into each iteration of the $P_{t\rightarrow t+1}$ process to provide an effective motion priori to each update of $O_{t,n}$.

\subsection{Splatting-based Alignment Method} \label{subsection:splat}

The motion feature extracted from the frame $t-1$ cannot directly be input into the frame $t$'s OFE because its coordinate system is not aligned with the frame $t$. On the contrary, if the motion feature comes from the frame $t$'s backward flow estimation process, this problem will not be encountered, but it will multiply the computational costs. 
We introduce the differentiable forward warping transformation, namely splatting, to efficiently align the motion feature from the video frame interpolation field \cite{niklaus2020softmax}.

Splatting is the inverse operation of sampling. Figure \ref{fig:splatting} (a) and (b) show the relationship between the two. The four rectangles in Fig. \ref{fig:splatting} (a) represent adjacent integer pixels in an image. Their pixel values are $V_1$, $V_2$, $V_3$, and $V_4$. The circle in the middle represents the sub-pixel to be sampled, and its pixel value is $V_0$.

$V_0$ is expressed as the weighted sum of adjacent four integer pixel values when bi-linear sampling is used. The mathematical expression is as follows:
\begin{align}
    & V_0 = \sum_{i = 1}^{4} w_i V_i, \label{eq:sampling_0} \\
    & w_i = (1 - \Delta x_i) (1 - \Delta y_i), \label{eq:sampling_01} \\
    & \Delta x_i = |x_i - x_0|, \label{eq:sampling_02} \\
    & \Delta y_i = |y_i - y_0|, \label{eq:sampling_03}
\end{align}
where $x_i$ and $y_i$ are the x and y coordinates of the pixel $i$.

From Eq. (\ref{eq:sampling_01}), (\ref{eq:sampling_02}), and (\ref{eq:sampling_03}), four weights have the following relationship:
\begin{equation}
    \sum_{i = 1}^{4} w_i = 1.
\end{equation}
Therefore, Eq. (\ref{eq:sampling_0}) is written in the following form:
\begin{align}
    & V_0 = \sum_{i = 1}^{4} n_i V_i \label{eq:sampling_1}, \\
    & n_i = \frac{w_i}{\sum_{i = 1}^{4} w_i} = w_i. \label{eq:sampling_11}
\end{align}

From Eq. (\ref{eq:sampling_1}) and (\ref{eq:sampling_11}), it can be concluded that when sampling a sub-pixel, one should compute the normalized weighted sum of all adjacent integer pixel values.
The inverse of this conclusion is splatting; that is when splatting an integer pixel, one should compute the normalized weighted sum of all adjacent sub-pixel values (Note that the first splatting proposed by Wang \etal \cite{wang2018occlusion} uses a non-normalized weighted sum, which leads to brightness inconsistencies \cite{niklaus2020softmax}). The mathematical expression is as follows:
\begin{align}
    & V_0 = \sum_{i = 1}^{k} n_i V_i \label{eq:splatting_0}, \\
    & n_i = \frac{f(w_i)}{\sum_{i = 1}^{k} f(w_i)}, \label{eq:splatting_01} \\
    & w_i = \text{max}(0, 1 - \Delta x_i) \cdot \text{max}(0, 1 - \Delta y_i) \label{eq:splatting_02}.
\end{align}
Here, $k$ is the number of all sub-pixels in an image. We can guarantee that the contribution of all sub-pixels, whose distance exceeds 1 pixel, is zero based on the $\text{max}(\cdot)$ operation in Eq. (\ref{eq:splatting_02}). $f(\cdot)$ is a mapping operation to $w_i$.
Figure \ref{fig:splatting} (b) shows the schematic illustration of splatting, in which the middle rectangle represents the integer pixel to be splatted and the surrounding circles are all sub-pixels.

The differences in $f(\cdot)$s determine how the splatting algorithm handles conflicts that multiple sources map to the same target.
Different $f(\cdot)$s from commonly used splatting types, such as average splatting and softmax splatting \cite{niklaus2020softmax}, are listed in Eq. (\ref{eq:splatting_f_average}) and (\ref{eq:splatting_f_softmax}), respectively:
\begin{align}
    & f^{a}(w_i) = w_i \label{eq:splatting_f_average}, \\
    & f^{s}(w_i) = \text{exp}(Z_i) \cdot w_i \label{eq:splatting_f_softmax},
\end{align}
where $Z$ is a metric that measures the mapping priority of the source. Most intuitively, $Z$ can represent the inverse depth of the sub-pixel; that is, the bigger the depth, the more likely it will be occluded and covered by other sources. Unfortunately, depth is often challenging to obtain. An alternative approach \cite{niklaus2020softmax} is to use the feature similarity between source and target as $Z$. Softmax splatting can strengthen the mapping ability of the high-importance source and realize the translation invariance of importance compared with average splatting. As a result, it works best in theory to resolve conflicts.

We propose a new motion feature alignment method based on the splatting concept.
We use $O_{t-1,n}$ to splat $M_{t-1,n}$ to the frame $t$'s coordinate.
Each feature vector $V_i$ in $M_{t-1,n}$, through its optical flow $o_i$ in $O_{t-1,n}$, determines its sub-pixel position $p_i$ in the frame $t$.
Then, with $p_i$ as the center, $V_i$ distributes the non-normalized contribution $w_i V_i$ to the four adjacent integer pixels. Finally, all contributions distributed to the same integer pixel $S_j$ are normalized and added to obtain the aligned motion feature $A_{t-1,n}$. Figure \ref{fig:splatting} (c) depicts the schematic illustration of the proposed Splatting-based motion feature alignment method.

We design our basic model according to Eq. (\ref{eq:splatting_f_average}). We must also determine a $Z$ to use the concept of softmax splatting, \ie, Eq. (\ref{eq:splatting_f_softmax}). 
First, we use $O_{t-1,n}$ to bi-linear sample $F_t^i$ to get $F_{t \rightarrow t-1}^i$.
Then, we calculate the similarity between the same position feature vectors $f_i$ and $g_i$ in $F_{t-1}^i$ and $F_{t \rightarrow t-1}^i$ and obtain Z. The mathematical expression is as follows:
\begin{align}
    & s_{i} = \frac{(f_i)^T \cdot g_i}{\sqrt{D}}, \label{eq:splatting_21} \\
    & Z_i = \alpha \cdot s_{i} \label{eq:splatting_22},
\end{align}
where $D$ is the channel number of $f_i$ and $g_i$, and $\alpha$ is a learnable parameter that controls the enhancement amplitude of the high-importance source's mapping ability. 

In ablation studies, the performances of the two splatting types are similar. Considering softmax splatting requires an additional consumption to calculate the feature similarity and its exponential value between the source and target pixels, which leads to an increase in inference time, we choose average splatting as our final version.

\subsection{Final-to-All Embedding Method} \label{subsection:o2m}

Apart from the encoding and alignment, embedding the motion feature into the two-frame backbone network is the most critical aspect of the SplatFlow framework. 
The red part in Fig. \ref{fig:overview} shows the schematic diagram of the proposed Final-to-All embedding method. Here, Final-to-All refers to the following: the network selects the previous frame's final iteration's aligned motion feature (\textbf{Final}) and inputs it into every iteration (\textbf{All}) of the current frame's estimation process.

On the one hand, we select the final iteration's aligned motion feature $A_{t-1,n}$ to be embedded into the two-frame backbone network.
Having only the same additional input is essential to ensure that every iteration has a unified motion reference. This approach is crucial as it reduces the ambiguity of early iterations, making it more conducive to the stability and convergence of the estimation result. The reason for choosing the final iteration is that the estimation error of the final iteration is the lowest. In addition, it brings a speed advantage since the only aligned motion feature of the final iteration needs to be calculated instead of all iterations.

On the other hand, every iteration will have the aligned motion feature as an additional input without exception.
This process is advantageous because it allows the aligned motion feature to participate deeply in the network estimation process and meet the different needs of different iteration contexts for using the aligned motion feature.

The last concern is how to input the final iteration's aligned motion feature $A_{t-1,n}$ into each iteration during the current frame's OFE.
In the original GRU Predictor network, the input of the GRU Cell is the concatenation of the context feature $F^{c}_{t}$ and the dynamically updated motion feature. In contrast, we concatenate $A_{t-1,n}$ with the above two features as a new input of the GRU Cell in the proposed Final-to-All embedding method.

The concatenation operation allows the network to intelligently select from or combine the three features without prescribing exactly how it is to do this, as described in GMA \cite{GMA}. Regarding occlusions, the network will refer to the previous frame's aligned motion feature when the optical flow cannot be inferred according to the current context.

\subsection{Training and Inference} \label{subsection:training}

Our method for training involves using three consecutive frames as a data unit and the flows of the previous two frames to supervise the network's outputs. We supervise our network on the $l_1$ distance between the predicted and the ground truth flow over the full sequences of the previous two frames' predictions, $\{f_{(t-1,1)}, ..., f_{(t-1,n)}\}$ and $\{f_{(t,1)}, ..., f_{(t,n)}\}$, with exponentially increasing weights. Given the ground truth flow $g_{t-1}$ and $g_{t}$, the loss is defined as follows: 
\begin{align}
    L
    & = L_{t-1} + L_{t} \\
    & = \sum_{i=1}^{n} \gamma^{n - i} ||g_{t-1} - f_{t-1,i}||_1 + \sum_{i=1}^{n} \gamma^{n - i} ||g_{t} - f_{t,i}||_1,
\end{align}
where we set $\gamma=0.85$ in our experiments.
Since the KITTI \cite{KITTI} data does not have the ground truth of the first frame, \ie, $g_{t-1}$, we use Eq. (\ref{eq:loss_kitti}) to calculate their losses:
\begin{equation}
    L = L_{t} = \sum_{i=1}^{n} \gamma^{n - i} ||g_{t} - f_{t,i}||_1. \label{eq:loss_kitti}
\end{equation}

When inferring the image sequence, the network calculates the final iteration's aligned motion feature $A_{t,n}$ for each frame except for the sequence's first frame and embeds $A_{t,n}$ to the next frame's estimation process under the proposed SplatFlow Framework.
As for the inference of the first frame, it is consistent with the original two-frame backbone. That is, we directly reserve the two-frame backbone's GRU Predictor network for inferring the first frame of an image sequence.

\section{Experiments}

\input{table/dataset}

\subsection{Datasets and Experimental Setup}

\subsubsection{Datasets} \label{subsection:datasets}

Mainstream public optical flow datasets are crucial for developing and testing optical flow algorithms. The commonly used datasets include FlyingChairs \cite{Flownet_and_Chairs}, FlyingThings3D \cite{Things}, Sintel \cite{Sintel}, KITTI2015 \cite{KITTI}, and HD1K \cite{HD1K}. Except for FlyingChairs \cite{Flownet_and_Chairs}, other datasets include video sequences with frames equal to or more than 3.
Our method requires valuable information from the historical frames, and the data unit comprises three consecutive frames during training. Therefore, we adjust several mainstream datasets containing consecutive frames, serving as training datasets in our experiments. Table \ref{tab:dataset} provides the statistical results of these datasets' overall, two-frame, and multi-frame versions.

\paragraph{FlyingChairs \cite{Flownet_and_Chairs}} This paper calls it Chairs for short. This is a synthetic dataset of moving 2D chair images superimposed on natural background images. It has 22232 and 640 two-frame image pairs for training and testing, with no temporal links. Therefore, it cannot build multi-frame datasets for training and testing.

\paragraph{FlyingThings3D \cite{Things}} This paper calls it Things for short. This synthetic dataset relies on randomness and a large pool of rendering assets to generate orders of magnitude more data than the previous options. It consists of 2239 training sequences and 437 testing sequences. Each consists of 10 consecutive clean pass images, showing that each sequence can provide 9 two-frame image pairs. When only the left camera and clean pass are considered, the dataset can provide 40,302 two-frame image pairs. When constructing a multi-frame dataset, we take 8 data units composed of three consecutive frames from each sequence. This results in a total of 35,824 training data units. In addition, we take the middle two and three frames from each testing sequence to form two-frame and multi-frame testing datasets containing 874 data units each.

\paragraph{Sintel \cite{Sintel}} This synthetic dataset derives from an open-source 3D animated short film, and it has important features such as long sequences, large motions, specular reflections, motion blur, defocus blur, atmospheric effects, and so on. It contains 23 training sequences and 12 testing sequences. Most sequences contain 50 frames along with 49 ground truths. This also reflects the original intention of the dataset design: to encourage the development of methods that use longer sequences and integrate information over time. If we only consider clean pass, there are 1041 two-frame image pairs for training a two-frame network. After collecting data units of three consecutive frames, we obtained a multi-frame training dataset containing 1018 data units. The testing dataset contains 552 units as the official number, except that the first unit of each sequence only includes two frames. Other units are all three frames.

\paragraph{KITTI2015 \cite{KITTI}} This paper calls it KITTI for short. This realistic dataset is obtained by annotating 400 dynamic scenes from the KITTI raw data collection using detailed 3D CAD models for all vehicles in motion. Two adjacent frames containing the expensive ground truth are selected from each scene to form training and testing datasets with 200 two-frame data units each. We recall the previous frame of each frame containing the ground truth from the KITTI raw data collection and construct new training and testing datasets with the unchanged number of data units. It is more challenging than KITTI2012 \cite{KITTI2012} for the OFE task, which only contains static scenes.

\paragraph{HD1K \cite{HD1K}} This realistic dataset comprises many difficult light and weather situations, such as low light, lens flares, rain, and wet streets. It contains 36 sequences and 1083 frames in total. Therefore, we can build a two-frame training dataset containing 1047 image pairs or a multi-frame training dataset containing 1011 data units based on it. Consistent with the previous method \cite{RAFT, GMA}, we did not divide the testing dataset separately.

\subsubsection{Evaluation Metrics}

\input{table/ablation}

It is common to use metrics the end-point error (EPE) and the percentage of optical flow outliers (Fl) when evaluating OFE.

EPE can be further divided into EPE-all (EPE), EPE-noc, and EPE-occ, depending on the scope of the evaluation, which are used to evaluate the average pixel end-point error over all, non-occluded, and occluded ground truth pixel regions, respectively.

Similarly, Fl can be divided into Fl-all, Fl-bg, and Fl-fg, which are used to evaluate the percentage of outliers averaged over all, background, and foreground ground truth pixel regions. Notably, the metric Fl is used primarily to evaluate KITTI \cite{KITTI} benchmark. The benchmark considers a pixel correctly estimated if the flow end-point error is \textless 3px or \textless 5\%.

\subsubsection{Implemental Details}

We use PyTorch to implement our method, and the two-frame backbone network of our final version is GMA \cite{GMA}. Compared with RAFT \cite{RAFT}, GMA \cite{GMA} will additionally input the current frame's aggregated global motion feature calculated through the attention mechanism at each GRU iteration. We use average splatting as our differentiable forward warping transformation.
We are consistent with GMA \cite{GMA} regarding training and inference hyperparameters, including the number of GRU iterations, the AdamW \cite{AdamW} as an optimizer, and the One-cycle \cite{one-cycle} as the learning rate adjustment strategy. We use the TITAN RTX GPU for training and inference. Unless otherwise noted, we evaluate after 24 flow iterations on Things \cite{Things} and KITTI \cite{KITTI} and 32 on Sintel \cite{Sintel}.

\subsubsection{Training Processes}

Our training process comprises C+T, S-finetune, and K-finetune. They are derived from the GMA \cite{GMA}'s training processes and have undergone some adjustments.

\paragraph{C+T} The training datasets used in GMA \cite{GMA} are Chairs \cite{Flownet_and_Chairs} and Things \cite{Things}. However, we use only Things \cite{Things}, but it is divided into two stages. In the first stage, we load and freeze GMA \cite{GMA}'s pre-training weights of C+T and separately train the parameters of the GRU Predictor and Convex Upsampling networks. In the second stage, we release all frozen parameters and finetune all parameters. The batch size in both stages is 8. The learning rate is $1.25 \times 10^{-4}$, and the number of training steps is 100K.

\paragraph{S-finetune} In this process, we finetune the model obtained by C+T on a mixed dataset dominated by Sintel \cite{Sintel}. These datasets compose of Things \cite{Things}, Sintel \cite{Sintel}, KITTI \cite{KITTI}, and HD1K \cite{HD1K}. We use a batch size of 8 and set the learning rate to $1.25 \times 10^{-4}$. The number of training steps is 120K. We use the finetuned model to be submitted and evaluate the testing dataset on the Sintel \cite{Sintel} benchmark.

\paragraph{K-finetune} In this process, we finetune the model obtained by S-finetune on a KITTI \cite{KITTI} data-enhanced hybrid dataset composed of the same four datasets. The batch size is 6, the learning rate is $1.25 \times 10^{-4}$, and the number of training steps is 50K. We use the finetuned model to be submitted and evaluate the testing dataset on the KITTI \cite{KITTI} benchmark.

\subsection{Ablation Studies} \label{subsection:ablation}
We conduct a series of ablation experiments on our method from six aspects, which include alignment method, differentiability of transformation, splatting type, embedding method, the number of iterations, and the number of frames, to examine their contributions. We use the model after the C+T training process in all experiments and evaluate the Things test \cite{Things}, Sintel train \cite{Sintel}, and KITTI train \cite{KITTI} datasets. 
Table \ref{tab:ablation} presents the quantitative results. Figure \ref{fig:iter} and \ref{fig:frame_num} depict the visualization results regarding the number of iterations and frames, respectively.

\subsubsection{Motion Feature Alignment Method}

We compare three motion feature alignment methods, denoted as ``Backward Flow'', ``Backward Warp'', and ``Forward Warp''. 
Unlike the other two methods, the ``Backward Flow'' baseline directly takes the backward motion feature in the backward flow estimation process from the frame $t$ to the frame $t-1$ as the aligned motion feature. The ``Backward Warp'' baseline uses the backward flow to bi-linear sample the forward motion feature of the frame $t-1$ to achieve feature alignment. The ``Forward Warp'' is our proposed method.

The ``Alignment Method'' part of Table \ref{tab:ablation} reveals the quantitative results of the three methods. From the results, our method outperforms the other two baselines in 5 of 6 dataset metrics, proving the effectiveness of the forward warping transformation in aligning the motion feature. 
As for the reasons why the performance of ``Backward Warp'' is inferior to ``Forward Warp'', we attribute it to the following: There are certain newly appeared regions in the frame $t$ compared to the frame $t-1$, which are occluded during the calculation of the backward flow from the frame $t$ to the frame $t-1$. As a result, the backward flow of these newly appeared regions cannot properly sample the forward motion feature, leading to erroneous motion features propagating to the next frame. However, for ``Forward Warp'', splatting will not affect these newly appeared regions, resulting in more accurate motion feature alignment.

Since there is no need to calculate the backward flow, the inference speed of the proposed method is nearly twice that of the other two baselines. 
Notably, the inference speed of ``Backward Flow'' and ``Backward Warp'' is very close, which also illustrates that the time consumption of the backward warping transformation is essentially negligible compared to calculating the backward flow.

\subsubsection{Differentiability of Forward Warping Transformation}

Here we analyze the impact of forward warping transformation's differentiability on performance. We design two non-differentiable forward warping transformation types: ``Non-differentiable Nearest'' and ``Non-differentiable Splatting''. The former fills the motion feature to the nearest integer pixel of the target sub-pixel position. At the same time, the latter is constructed by disabling the gradient propagation of our method's splatting transformation during training.
From the ``Differentiability'' part of Table \ref{tab:ablation}, we find that our proposed ``Differentiable Splatting'' is better than ``Non-differentiable Splatting'', and ``Non-differentiable Splatting'' is better than ``Non-differentiable Nearest'' in all dataset metrics. It shows that both differentiability and sub-pixel level filling improve the performance of forward warping transformation on OFE.

\subsubsection{Splatting Type}

We design the ``Softmax Splatting'' baseline based on Eq. (\ref{eq:splatting_f_softmax}) to observe the impact of splatting type on the estimation effect. On the other hand, ``Average Splatting'' is our proposed method, based on Eq. (\ref{eq:splatting_f_average}). The ``Splatting Type'' part of Table \ref{tab:ablation} shows the quantitative results of the two methods.

Compared with ``Softmax Splatting'', ``Average Splatting'' can achieve considerable or even better results with the best performance in 4 of 6 dataset metrics, which shows that the average splatting can be better generalized to more scenes in OFE tasks. We also discover that ``Softmax Splatting'' has a weak advantage over ``Average Splatting'' in the Things val \cite{Things} Clean and Sintel train \cite{Sintel} Clean datasets. We believe that the texture of clean pass data is simpler than realistic and final pass data, making it easier to extract more discriminative image feature representation, which is conducive to softmax splatting, playing its theoretical advantages in eliminating mapping conflicts.
Furthermore, "Average Splatting" has a faster inference speed, as it does not need to calculate the feature similarity and its exponential value between source and target pixels.

\input{fig/embedder}

\subsubsection{Motion Feature Embedding Method}

We design a Final-to-All embedding method for the single-resolution iterative two-frame backbone networks. We also designe two intuitively reasonable baselines, the One-to-One and the Final-to-Final embedding methods, to prove the effectiveness of the proposed embedding method. Fig. \ref{fig:overview}'s red part and Fig. \ref{fig:embedder} show the structural differences between the three methods.

The One-to-One embedding method baseline uses the number of iterations as the selection basis to embed the different features in the sequence $S=\{A_{t-1,1},A_{t-1,2},...,A_{t-1,n}\}$ into the various GRU iterations of the frame $t$. Similar to the multi-frame method for pyramid backbones, this method is designed for information interaction within the same layer/iteration, which is not ideal for fully utilizing the unique ``single-resolution'' characteristic of the single-resolution iterative network.
In the Final-to-Final embedding method baseline, $A_{t-1,n}$ is directly embedded in the final GRU iteration of the frame $t$. The embedding method is designed from the perspective of flow post-processing, so it is a backbone-independent multi-frame method, which has the advantages of more flexibility and easy implementation.

The ``Embedding Method'' part of Table \ref{tab:ablation} shows that the proposed Final-to-All embedding method is superior to the One-to-One and the Final-to-Final methods in all dataset metrics. It proves that using the final iteration's aligned motion feature as a unified motion reference to participate deeply in the network estimation process can maximize the promoting effect of the multi-frame settings. At the same time, the inference speed of the Final-to-All embedding method is comparable to that of the Final-to-Final, indicating that the time consumption of the aligned motion feature as an additional input to the GRU Cell is minimal. 
Additionally, the Final-to-All embedding method has the apparent speed advantage over One-to-One due to the feature selection rule that only calculates the aligned motion feature for the final, rather than every, iteration.

\subsubsection{Number of Iterations}

\input{fig/iter}

To make a fair comparison with the previous methods, we adopt a large number of iterations for inference when evaluating the datasets (24 for Things \cite{Things} and KITTI \cite{KITTI}, 32 for Sintel \cite{Sintel}). Like other single-resolution iterative methods, our method quickly converges, comparable with final results only after 12 iterations, as shown in Fig. \ref{fig:iter}. Notably, our method surpasses GMA \cite{GMA}'s final performance in less than five average iterations. When we compare the final time consumed by GMA \cite{GMA} on Sintel \cite{Sintel} (32 iterations), we see an increase in speed of 4.01 times, \ie, 5.61 fps vs. 22.49 fps on the TITAN RTX GPU.

This experiment demonstrates that our proposed method is highly adapted to the ``iterative'' characteristic of backbones and has significant advantages in terms of performance.

\subsubsection{Number of Frames}

\input{table/multi-frame}

We extend our SplatFlow to cases with more than three frames to explore the impact of more frames' information on the network's performance.
To achieve this aim, we warp the motion feature of any historical frame, such as the frame $t-k$'s $M_{t-k}$, through intermediate frames' flows to the current frame. This process obtains $A_{t-k \rightarrow t}$, the motion feature from the frame $t-k$ that is aligned to the frame $t$. The mathematical expression is as follows:
\begin{equation}
    A_{t-k\rightarrow t}=\operatorname{Splat}(...\operatorname{Splat}(M_{t-k,n},O_{t-k,n})...,O_{t-1,n}).
\end{equation}
Next, the feature set $\{ A_{t-k \rightarrow t}, A_{t-k+1 \rightarrow t}, ..., A_{t-1 \rightarrow t}\}$ is concatenated and input into the estimation process of the current frame.

\input{fig/frame_num}

We adjust the Things, Sintel, and KITTI datasets to 4-frame and 5-frame versions following Section \ref{subsection:datasets}. We evaluate the 4-frame version model and 5-frame version model on the Things, Sintel, and KITTI datasets after the C+T training process. The results are shown in Fig. \ref{fig:frame_num}.

We find that all our multi-frame models ($\geq$3) outperform the two-frame backbone GMA \cite{GMA} on Things \cite{Things} and KITTI \cite{KITTI}. This further proves that introducing the multi-frame settings is beneficial for improving the optical flow estimation accuracy. At the same time, we find that introducing more frames ($\textgreater$3) can decrease the performance of Sintel \cite{Sintel} to worse than GMA \cite{GMA}. This result arises because the models cannot be well generalized to Sintel \cite{Sintel} with dramatic changes in motion when trained on Things with small changes in motion.

Finally, we find that our 3-frame model outperforms all other models on all datasets. We think this is because the 3-frame settings can already provide sufficient and reliable multi-frame temporal context, and introducing more frames will aggravate the accumulation and diffusion of errors from the historical frames’ estimations. The above indicates that the 3-frame model possesses superior learning and generalization capabilities.

\subsection{Comparisons with Multi-frame Baselines}

In this subsection, we compare the proposed method with other multi-frame methods to show and analyze the advantages of our method.

The multi-frame methods are divided into backbone-dependent and backbone-independent. We use the initialization method called WarmStart proposed in RAFT \cite{RAFT} as a backbone-dependent multi-frame baseline based on single-resolution iterative backbones. Additionally, we choose FlowFusion \cite{ren2019fusion}, which has shown promising results on PWC-Net \cite{PWCNet}, as the backbone-independent multi-frame baseline. We use three methods on RAFT \cite{RAFT} and GMA \cite{GMA} backbones; the quantitative results are shown in Table \ref{tab:multi-frame}.

First, our method has proven superior to FlowFusion \cite{ren2019fusion} in all dataset metrics, regardless of the backbone used. It shows that fully utilizing the backbone network's ``single-resolution'' and ``iterative'' characteristics could achieve better estimation performance. Additionally, our inference speeds are nearly three times that of FlowFusion \cite{ren2019fusion}. This speed advantage is attributed to the splatting transformation, which can continuously transmit the motion feature in one direction, thus avoiding the time consumption of calculating the backward flow. If we only consider inferring non-first video frames, our methods would require fewer parameter amounts than FlowFusion \cite{ren2019fusion}.

Second, compared with WarmStart \cite{RAFT}, our methods perform best in 5 of 6 dataset metrics, demonstrating its ability to better use the multi-frame temporal context to improve performance effectively. The ``GMA+WS'' row in Fig. \ref{fig:occ} (a)-(d) shows the qualitative results of the WarmStart \cite{RAFT} method on the Sintel \cite{Sintel} test Clean dataset. It can be seen from Fig. \ref{fig:occ} that although ``GMA+WS'' has made slight improvement compared with ``GMA'', it still struggles to cope with the challenging occlusion problems correctly. In many cases, the ``GMA+WS'' performance is unchanged or decreased compared with that of the backbone, as reflected in the performance of ``GMA+WarmStart'' on the Things \cite{Things} val dataset in Table \ref{tab:multi-frame}. Furthermore, the ``forward\_interpolate'' operation used in WarmStart \cite{RAFT} is on the CPU and therefore takes longer to make inferences.

\input{table/multi-occ}

\input{table/two-frame}

\input{fig/occ}

\input{table/occ}

\input{table/benchmarks}

\input{table/final}

Furthermore, we evaluate the performance of SplatFlow and the other two methods in occluded regions, as shown in Table \ref{tab:multi-occ}. We find that all methods can improve their performance in occluded regions compared with backbone networks, proving that the multi-frame settings can indeed help to deal with the occlusion problem. Compared to the other two baselines, our method improves significantly in occluded regions, further demonstrating that our method can make more effective use of multi-frame temporal context to solve occlusions. At the same time, the decreased performance of WarmStart \cite{RAFT} in non-occluded regions also indicates that the multi-frame settings have advantages and disadvantages. Leveraging its strengths and avoiding its weaknesses is a direction worth further in-depth research.

In sum, the above experiments demonstrate that our method has comprehensive and significant advantages in using multi-frame information with good speed and accuracy.

\subsection{Comparisons with Two-frame Backbone}

In this subsection, we compare the proposed method with the single-resolution backbone networks to analyze the role of the multi-frame settings.

First, we take RAFT \cite{RAFT} and GMA \cite{GMA} as backbone networks, showing their quantitative results and multi-frame versions after the C+T training process. As shown in Table \ref{tab:two-frame}, the networks can achieve significant performance improvements in almost all dataset metrics after embedding the previous frame's motion feature compared with the original backbone networks. At the same time, the two multi-frame versions maintain similar inference speeds with their backbones due to the unidirectional motion feature aggregation method based on splatting transformation. If we only consider inferring non-first video frames, the multi-frame versions would have comparable parameter amounts with their backbones.

It is worth noting that the ``SplatFlow-GMA'' baseline, \ie, our final model, as ``RAFT'' baseline considering both the multi-frame temporal and spatial contexts \cite{GMA}, outperforms the ``SplatFlow-RAFT'' baseline or ``GMA'' baseline, as can be seen in Table \ref{tab:two-frame}. This result shows that the multi-frame temporal and spatial contexts are two complementary contexts that cannot be replaced by each other. Moreover, it further illustrates that the multi-frame temporal context is indispensable to improving OFE accuracy. Additionally, the ``SplatFlow-RAFT'' baseline outperforms the ``GMA'' baseline on inference speed and 4 of 6 dataset metrics, and parameter amount if only considering inferring non-first video frames. The finding shows that, in most cases, the multi-frame temporal context is even more valuable than the spatial context.

We further explore the impact of the multi-frame settings on occlusions. Table \ref{tab:occ} shows the evaluation results and relative improvements compared ``SplatFlow-RAFT'' and ``SplatFlow-GMA'' baselines with their two-frame backbones RAFT \cite{RAFT} and GMA \cite{GMA} on three types of regions (non-occluded, occluded, and all) on Things \cite{Things} val and Sintel \cite{Sintel} train Clean datasets after the C+T training process and on Sintel train and Sintel test datasets after the S-finetune training process. 
We exclude the KITTI \cite{KITTI} dataset from evaluation, considering that the KITTI \cite{KITTI} dataset has no ground truth of occluded regions. 
From the results, our methods achieve significant improvements in all three regions of all datasets after all training processes. 
However, the improvements in occluded regions are the most significant, showing that the network can benefit from the multi-frame settings in every region, especially in occluded regions. 

Figure \ref{fig:occ} shows the qualitative results of our method and GMA on the Sintel test Clean dataset after the S-finetune and the KITTI test dataset after the K-finetune training processes. The solid-box marked regions are significantly occluded in the frame $t+1$, and the dotted-box marked regions are non-occluded but challenging to estimate. The box contents show that our method can obtain more detailed results in non-occluded regions while achieving more acceptable performance in occluded regions, avoiding the large area of failure. Meanwhile, the evaluation values reported on the Sintel benchmark in Fig. \ref{fig:occ} (a)-(c) show that our method has surpassed GMA in all three regions, consistent with the conclusion in Table \ref{tab:occ}.

These results show that the proposed method effectively uses the multi-frame settings to optimize the effect of OFE comprehensively.

\subsection{Results on Benchmarks}

We evaluate our method on public Sintel \cite{Sintel} and KITTI \cite{KITTI} benchmarks and compare the results against previous works, shown in Table \ref{tab:benchmarks}.

First, after the C+T training process (first part in Table \ref{tab:benchmarks}), we significantly exceed all the existing methods on most of the metrics on the Sintel \cite{Sintel} train and KITTI \cite{KITTI} train datasets, demonstrating our method's superior performance on cross-dataset generalization.

Second, after the S-finetune training process (second part in Table \ref{tab:benchmarks}), our method ranks first on both the Sintel \cite{Sintel} test Clean and Sintel \cite{Sintel} test  Final datasets, with EPEs of 1.12 and 2.07, a 19.4\% and 16.2 \% decrease in error compared to the previous best method GMA \cite{GMA}. At the same time, we also achieve the lowest error on the Sintel \cite{Sintel} train Clean and Final datasets, which are the components of the training datasets during the S-finetune process. This result proves that our method can dig and use useful clues at a deeper level, thus achieving a better training convergence effect than the others.

Finally, after the K-finetune training process (third part in Table \ref{tab:benchmarks}), our method ranks first among all methods based on optical flow on the KITTI \cite{KITTI} test dataset. It is worth noting that our performance on the training datasets of this process is not outstanding. It is because we reduce the proportion of KITTI \cite{KITTI} data compared with other methods during training. This reduction prevents overfitting the KITTI \cite{KITTI} train dataset, which only has a small amount of data.

From these results, it can be seen our method has a higher generalization performance than the previous best method and has reached a new state-of-the-art performance on two public benchmarks, thus proving its effectiveness and advancement.

\input{fig/final_error}

\subsection{Discussion}

This subsection explores why the proposed method performs slightly worse than its backbone on the Sintel \cite{Sintel} train Final dataset in Table \ref{tab:two-frame}. Considering that the appearances and motion distributions of objects in the same video sequence are similar, we evaluate the two methods sequence by sequence on the Sintel \cite{Sintel} train Final dataset to find video sequences challenging to evaluate. Table \ref{tab:final} shows the quantitative results.

A total of 23 video sequences are in Table \ref{tab:final}. The error increment of the ``ambush\_4'' sequence is extremely large, while other sequences' are negative or small enough to be ignored.

Through careful observation of the ``ambush\_4'' sequence, we find that our method has large errors primarily in two cases; example qualitative results are shown in Fig. \ref{fig:final_error}. In the first case, the image is seriously blurred, and the motion displacement is enormous; it cannot be judged even by human eyes, as shown in Fig. \ref{fig:final_error} (a). In this case, both methods fail and produce significant errors. Therefore the relative error is also large and random. In the other case, the image has large blurs that are not found in the Things \cite{Things} train dataset during the C+T training process, as shown in Fig. \ref{fig:final_error} (b). In this case, our method will produce fog-like error estimation in fuzzy areas, which does not exist in the Sintel \cite{Sintel} test Final dataset after the S-finetune training process. This result shows that our method has poor generalization for fuzzy texture compared with GMA \cite{GMA}. This issue can be improved using data enhancements that simulate blurring or by adding training samples with blurring. 

\section{Conclusion}

In summary, we propose a MOFE framework named SplatFlow. SplatFlow introduces the differentiable splatting transformation to align the motion feature and designs a Final-to-All embedding method to input the aligned motion feature into the current frame's estimation, thus remodeling the existing two-frame backbones. 
The proposed SplatFlow framework has significant occlusion coping ability and balanced evaluation accuracy and speed.

Through this work, we further prove that the multi-frame settings play active and indispensable roles in solving occlusions and improving OFE accuracy. It will be a potential research direction to explore more effective multi-frame methods.

\input{body/tail}

\end{document}

%% file: body/config.tex
\smartqed  
\usepackage[usenames]{color}
\usepackage[numbers]{natbib}
\usepackage[table]{xcolor}
\usepackage{amssymb}
\usepackage{makecell}
\usepackage{multirow}
\usepackage{rotating}
\usepackage{array}
\usepackage{times}
\usepackage{graphicx}
 \usepackage{psfrag}
\usepackage{subfigure}
\usepackage{enumitem}
\usepackage{rotating}
\usepackage{url}
\usepackage{rotating, graphicx}
\usepackage{amsmath}
\usepackage{booktabs}
\usepackage{lscape}
\usepackage{verbatim}
\usepackage{geometry}
\usepackage[misc]{ifsym}
\papertype{generic article}
\usepackage[normalem]{ulem}
\useunder{\uline}{\ul}{}

\newcommand{\etal}{\textit{et al}.}
\newcommand{\ie}{\textit{i}.\textit{e}.}
\newcommand{\eg}{\textit{e}.\textit{g}.}

\usepackage[pagebackref=true,breaklinks=true,linkcolor=red,anchorcolor=blue, citecolor=green,letterpaper=true,colorlinks,bookmarks=false]{hyperref}
\usepackage{breakurl}
\usepackage{graphicx}
\definecolor{DarkBlue}{rgb}{0,0,1}
\definecolor{DarkRed}{rgb}{0.6,0.00,0.08}
\definecolor{DarkGreen}{rgb}{0.0,0.6,0.08}
\definecolor{LightBlue}{rgb}{0.88,0.92,0.95}
\definecolor{Orange}{rgb}{1,0.5,0}
\UseRawInputEncoding

%% file: body/head.tex
\title{SplatFlow: Learning Multi-frame Optical Flow via Splatting}
\subtitle{}

\author{Bo Wang $^{1}$  \and
        Yifan Zhang $^{1}$ \and
        Jian Li $^{1}$ \and \\
        Yang Yu $^{1}$ \and
        Zhenping Sun $^{1}$ \and
        Li Liu $^{2}$ \and
        Dewen Hu $^{1}$
}

\authorrunning{Bo Wang \etal} 

\institute{
Bo Wang (wb@nudt.edu.cn) \\
Yifan Zhang (zhangyifanedu@nudt.edu.cn) \\
Jian Li (lijian@nudt.edu.cn) \\
Yang Yu (yuyangnudt@hotmail.com) \\
Zhenping Sun (sunzhenping@nudt.edu.cn) \\
Li Liu (liuli\_nudt@nudt.edu.cn) \\
\textrm{\Letter} Dewen Hu (dwhu@nudt.edu.cn) \\
$^{1}$ College of Intelligence Science and Technology, National University of Defense Technology, Changsha, China. \\
$^{2}$ College of Electronic Science and Technology, National University of Defense Technology, Changsha, China.
}
\date{Received: 27 October 2022 / Accepted: 2 January 2024}

\maketitle

%% file: body/abstract.tex
\begin{abstract}
The occlusion problem remains a crucial challenge in optical flow estimation (OFE). Despite the recent significant progress brought about by deep learning, most existing deep learning OFE methods still struggle to handle occlusions; in particular, those based on two frames cannot correctly handle occlusions because occluded regions have no visual correspondences.
However, there is still hope in multi-frame settings, which can potentially mitigate the occlusion issue in OFE. Unfortunately, multi-frame OFE (MOFE)  remains underexplored, and the limited studies on it are mainly specially designed for pyramid backbones or else obtain the aligned previous frame's features, such as correlation volume and optical flow, through time-consuming backward flow calculation or non-differentiable forward warping transformation. 
This study proposes an efficient MOFE framework named SplatFlow to address these shortcomings. SplatFlow introduces the differentiable splatting transformation to align the previous frame's motion feature and designs a Final-to-All embedding method to input the aligned motion feature into the current frame's estimation, thus remodeling the existing two-frame backbones.
The proposed SplatFlow is efficient yet more accurate, as it can handle occlusions properly. Extensive experimental evaluations show that SplatFlow substantially outperforms all published methods on the KITTI2015 and Sintel benchmarks. 
Especially on the Sintel benchmark, SplatFlow achieves errors of 1.12 (clean pass) and 2.07 (final pass), with surprisingly significant 19.4\% and 16.2\% error reductions, respectively, from the previous best results submitted. The code for SplatFlow is available at \href{https://github.com/wwsource/SplatFlow}{https://github.com/wwsource/SplatFlow}.


\end{abstract}

%% file: fig/overview.tex
\begin{figure*}[t]
	\begin{center}
		\includegraphics[width=0.95\linewidth]{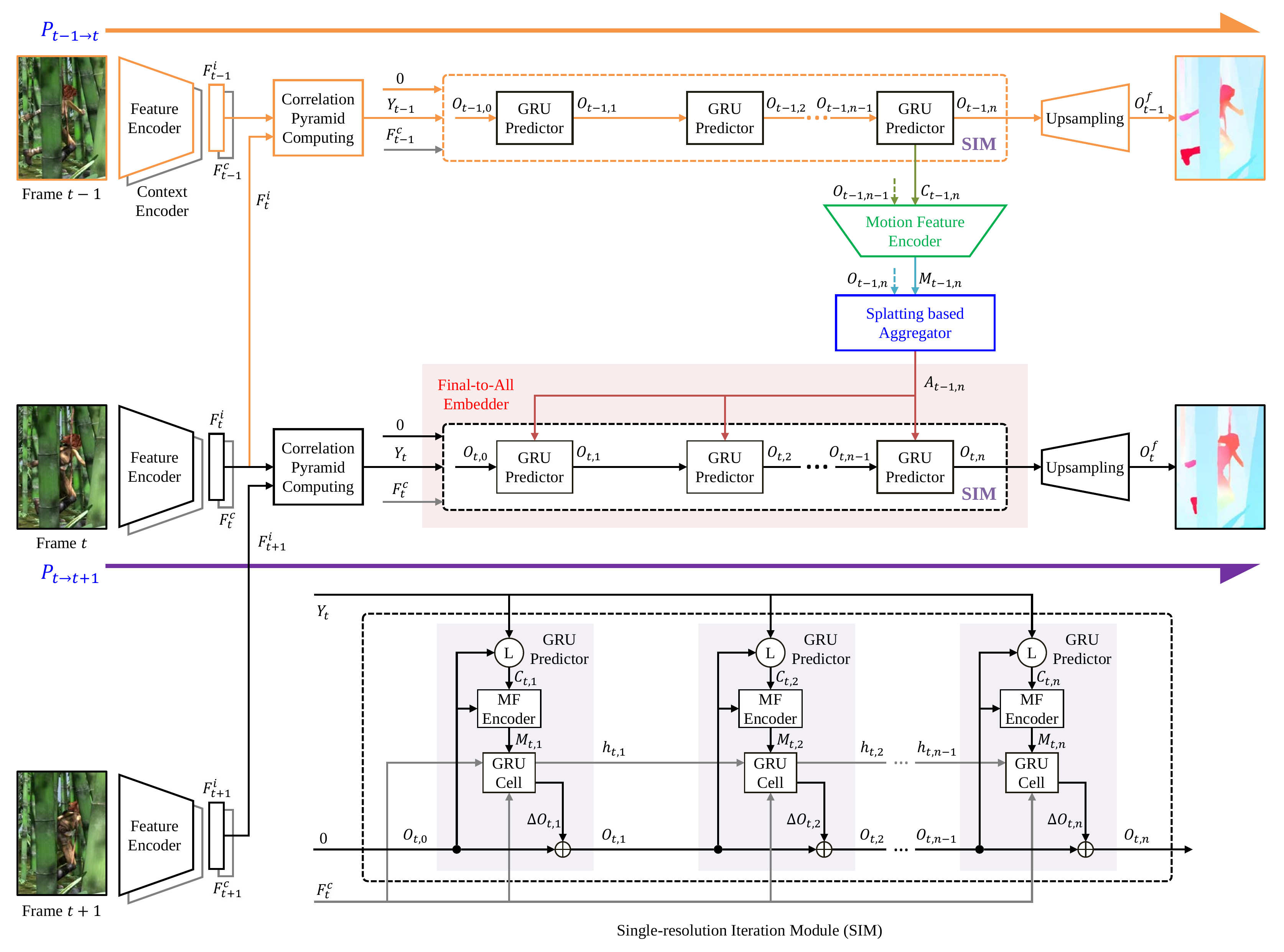}
	\end{center}
	\caption{
	The overall architecture of the proposed SplatFlow framework is designed for the single-resolution iterative backbone, \eg, RAFT \cite{RAFT}. 
	The thick orange and purple arrows show the estimation processes $P_{t-1\rightarrow t}$ and $P_{t\rightarrow t+1}$, respectively.
	The framework encompasses the encoding, alignment, and embedding of the motion feature. The encoder network encodes the motion feature (green part). We use a Splatting-based method to achieve the motion feature's unidirectional and differentiable alignment (blue part). We use a Final-to-All embedding method to input the aligned motion feature into the frame $t$'s estimation process (red part).
	The detailed data flow for iteratively updating optical flow in RAFT \cite{RAFT} is summarized as a single-resolution iteration module (SIM) in the purple part.
	}
	\label{fig:overview}
\end{figure*}

%% file: fig/mf.tex
\begin{figure}[h]
	\begin{center}
		\includegraphics[width=0.75\linewidth]{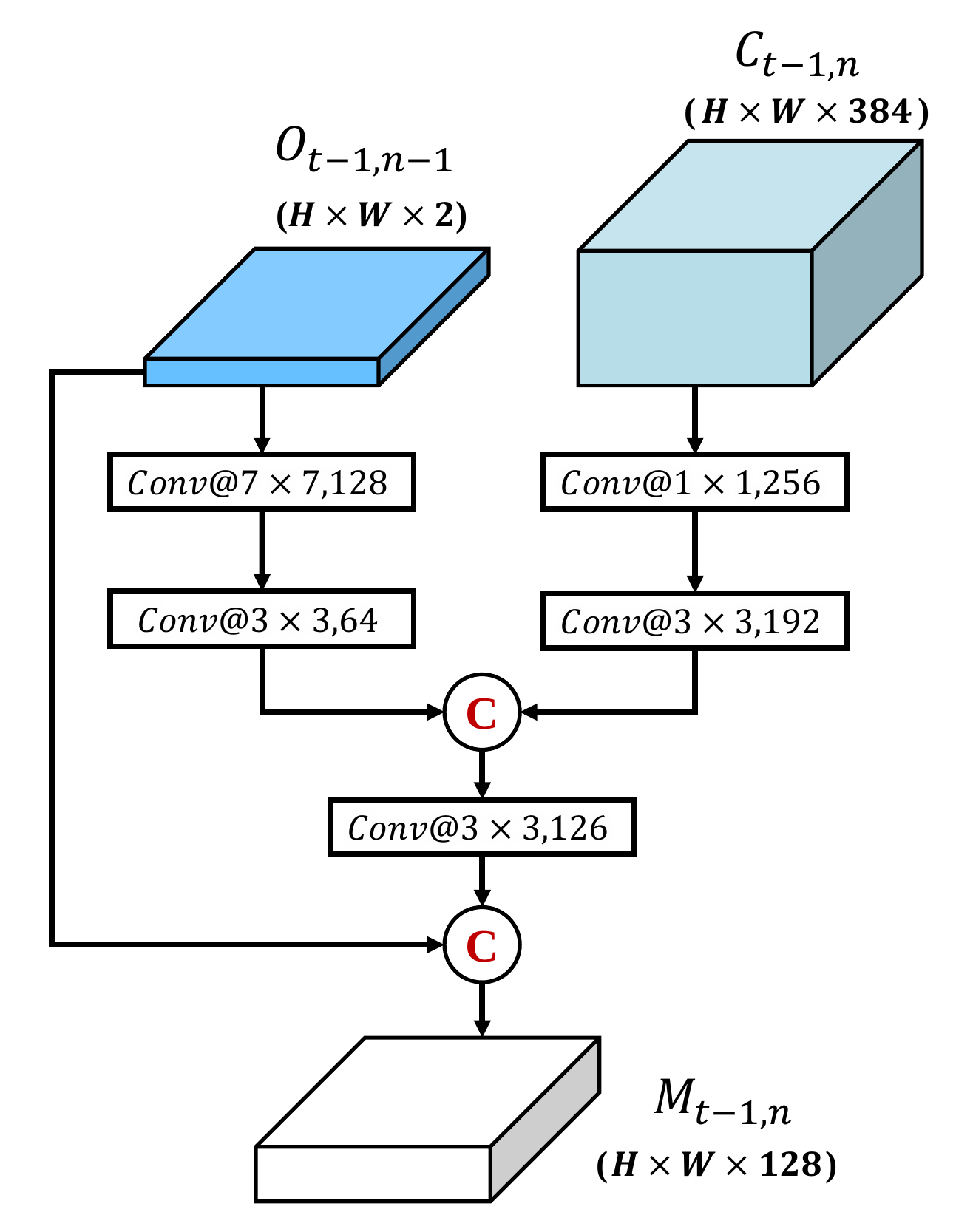}
	\end{center}
	\caption{
	Schematic illustration of the whole generation process of the motion feature $M_{t-1,n}$. $O_{t-1,n-1}$ is the frame $t-1$'s optical flow of the ${n-1}^{th}$ iteration, and $C_{t-1,n}$ is the correlation feature. We use $conv@n\times n,l$ to represent a convolution operation with a kernel of $n$, output channels of $l$, and a stride of 1. We use \textbf{\textcolor{red}{C}} to represent the concatenation operation.
	}
	\label{fig:mf}
\end{figure}

%% file: fig/splatting.tex
\begin{figure*}[t]
	\begin{center}
		\includegraphics[width=\linewidth]{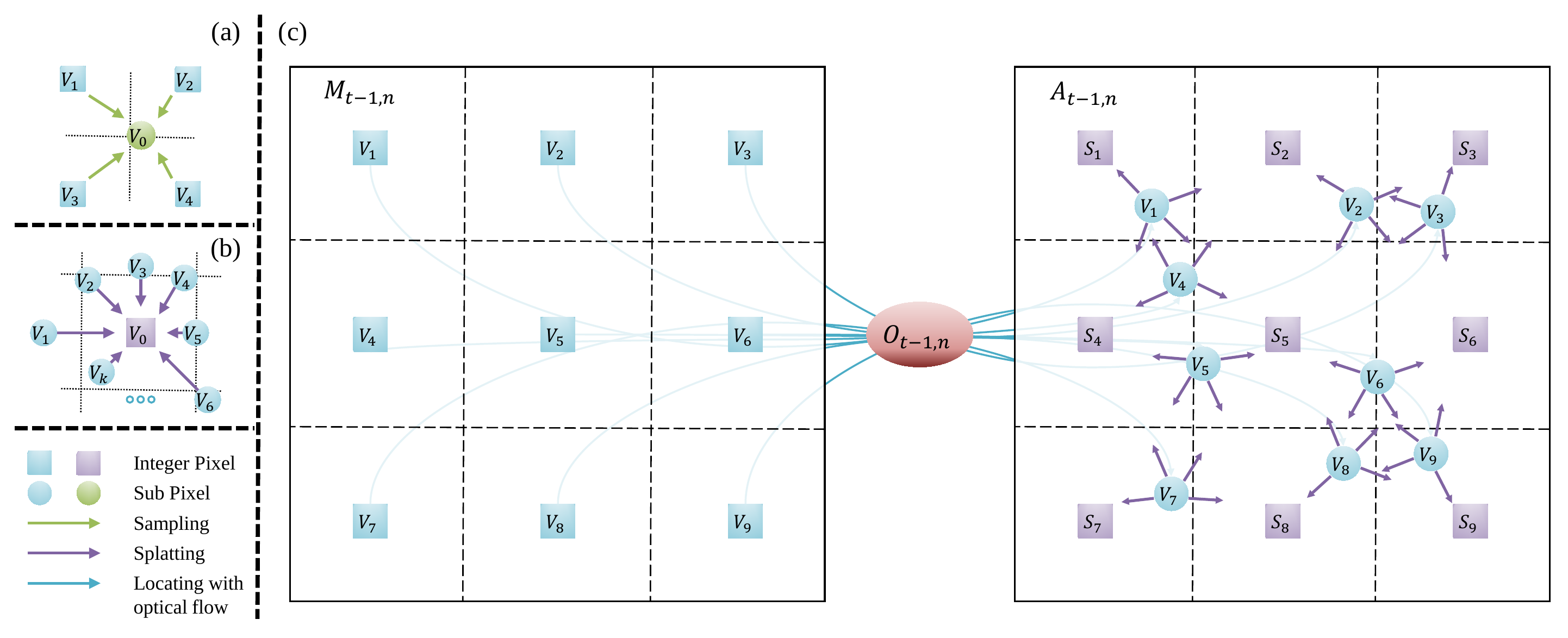}
	\end{center}
	\caption{
 (a) Schematic illustration of sampling. 
 (b) Schematic illustration of splatting.
 Blue rectangles/circles represent the pixels providing contributions in sampling or splatting. Green circles represent the pixels receiving contributions in sampling. Purple rectangles represent the pixels receiving contributions in splatting.
 The difference between (a) and (b) is that sampling uses the values of surrounding integer pixels to calculate the value of the sub-pixel, while splatting uses the values of surrounding sub-pixels to calculate the value of the integer pixel.
(c) Schematic illustration of the proposed Splatting-based motion feature alignment method. We use the frame $t-1$'s optical flow $O_{t-1,n}$ after the $n$th iteration to splat the motion feature $M_{t-1,n}$ to the frame $t$'s coordinate with non-normalized contributions. All contributions distributed to the same integer pixel $S_j$ are normalized and added to obtain the aligned motion feature $A_{t-1,n}$.
	}
	\label{fig:splatting}
\end{figure*}

%% file: table/dataset.tex
\begin{table*}[t]
\centering
\caption{
    Statistical results of the overall, two-frame, and multi-frame versions of the mainstream public optical flow datasets. Only the left camera and clean pass are considered for Things \cite{Things}, and only the clean pass is considered for Sintel \cite{Sintel}.
}
\label{tab:dataset}
\resizebox{0.85\textwidth}{!}{%
\begin{tabular}{ccccccccc}
\toprule
Dataset & \begin{tabular}[c]{@{}c@{}}Max.\\ Resolution\end{tabular} & \begin{tabular}[c]{@{}c@{}}Training\\ Scenes\\      Number\end{tabular} & \begin{tabular}[c]{@{}c@{}}Testing\\ Scenes\\      Number\end{tabular} & \begin{tabular}[c]{@{}c@{}}Average\\ Scene\\      Length\end{tabular} & \begin{tabular}[c]{@{}c@{}}Two-frame\\      Training Units\\      Number\end{tabular} & \begin{tabular}[c]{@{}c@{}}Multi-frame\\      Training Units\\      Number\end{tabular} & \begin{tabular}[c]{@{}c@{}}Two-frame\\      Testing Units\\      Number\end{tabular} & \begin{tabular}[c]{@{}c@{}}Multi-frame\\      Testing Units\\      Number\end{tabular} \\ \midrule
Chairs \cite{Flownet_and_Chairs} & 512$\times$384 & 22232 & 640 & 2 & 22232 & 0 & 640 & 0 \\
Things \cite{Things} & 960$\times$540 & 2239 & 437 & 10 & 40302 & 35824 & 874 & 874 \\
Sintel \cite{Sintel} & 1024$\times$436 & 23 & 12 & 46.51 & 1041 & 1018 & 552 & 552 \\
KITTI \cite{KITTI} & 1240$\times$380 & 200 & 200 & 21 & 200 & 200 & 200 & 200 \\
HD1K \cite{HD1K} & 2560$\times$1080 & 36 & 0 & 30.08 & 1047 & 1011 & 0 & 0 \\ \bottomrule
\end{tabular}%
}
\end{table*}

%% file: table/ablation.tex
\begin{table*}[t]
\centering
\caption{Quantitative results of ablation experiments. All data come from the C+T training process. The data marked with parentheses under the ``Parameters'' column indicates the model parameter amount used to infer non-first video frames.}
\label{tab:ablation}
\resizebox{0.8\textwidth}{!}{%

\begin{tabular}{cccccccccc}
\toprule
\multirow{3}{*}{Experiment} & \multirow{3}{*}{Variations} & \multicolumn{2}{c}{Things \cite{Things}} & \multicolumn{2}{c}{Sintel \cite{Sintel}} & \multicolumn{2}{c}{KITTI \cite{KITTI}} & \multirow{3}{*}{FPS} & \multirow{3}{*}{Parameters} \\
 &  & \multicolumn{2}{c}{val} & \multicolumn{2}{c}{train} & \multicolumn{2}{c}{train} &  &  \\ \cmidrule(lr){3-4} \cmidrule(lr){5-6} \cmidrule(lr){7-8}
 &  & Clean & Final & Clean & Final & EPE & Fl-all &  &  \\ \midrule
GMA \cite{GMA} & - & 4.98 & 4.23 & 1.31 & 2.75 & 4.48 & 16.9 & 12.96 & 5.9M \\ \midrule
\multirow{3}{*}{\begin{tabular}[c]{@{}c@{}}Alignment\\ Method\end{tabular}} & Backward Flow & 3.27 & 3.28 & 1.22 & \textbf{2.76} & 4.30 & 16.1 & 6.40 & 9.0M/(6.3M) \\
 & Backward Warp & 3.23 & 3.96 & 1.31 & 2.98 & 3.82 & 15.1 & 6.35 & 9.0M/(6.3M) \\
 & {\ul Forward Warp} & \textbf{3.02} & \textbf{3.23} & \textbf{1.22} & 2.97 & \textbf{3.70} & \textbf{14.9} & \textbf{12.29} & \textbf{9.0M/(6.3M)} \\ \midrule
\multirow{3}{*}{\begin{tabular}[c]{@{}c@{}}Differentiability\end{tabular}} & Non-differentiable Nearest & 4.11 & 4.62 & 1.27 & 3.15 & 4.56 & 17.0 & 12.29 & 9.0M/(6.3M) \\
 & Non-differentiable Splatting & 3.25 & 3.44 & 1.26 & 3.05 & 3.98 & 15.9 & 12.29 & 9.0M/(6.3M) \\
 & {\ul Differentiable Splatting} & \textbf{3.02} & \textbf{3.23} & \textbf{1.22} & \textbf{2.97} & \textbf{3.70} & \textbf{14.9} & \textbf{12.29} & \textbf{9.0M/(6.3M)} \\ \midrule
\multirow{2}{*}{\begin{tabular}[c]{@{}c@{}}Splatting\\ Type\end{tabular}} & Softmax Splatting & \textbf{2.97} & 3.30 & \textbf{1.20} & 2.97 & 3.92 & 15.2 & 11.97 & 9.0M/(6.3M) \\
 & {\ul Average Splatting} & 3.02 & \textbf{3.23} & 1.22 & \textbf{2.97} & \textbf{3.70} & \textbf{14.9} & \textbf{12.29} & \textbf{9.0M/(6.3M)} \\ \midrule
 \multirow{3}{*}{\begin{tabular}[c]{@{}c@{}}Embedding\\      Method\end{tabular}} & One-to-One & 3.45 & 3.43 & 1.27 & 3.10 & 4.20 & 16.2 & 10.56 & 9.0M/(6.3M) \\
 & Final-to-Final & 3.49 & 4.09 & 1.37 & 3.10 & 4.19 & 16.2 & \textbf{12.51} & 9.0M/(6.3M) \\
 & {\ul Final-to-All} & \textbf{3.02} & \textbf{3.23} & \textbf{1.22} & \textbf{2.97} & \textbf{3.70} & \textbf{14.9} & 12.29 & \textbf{9.0M/(6.3M)} \\ \bottomrule
\end{tabular}

}
\end{table*}

%% file: fig/embedder.tex
\begin{figure}[h]
	\begin{center}
		\includegraphics[width=\linewidth]{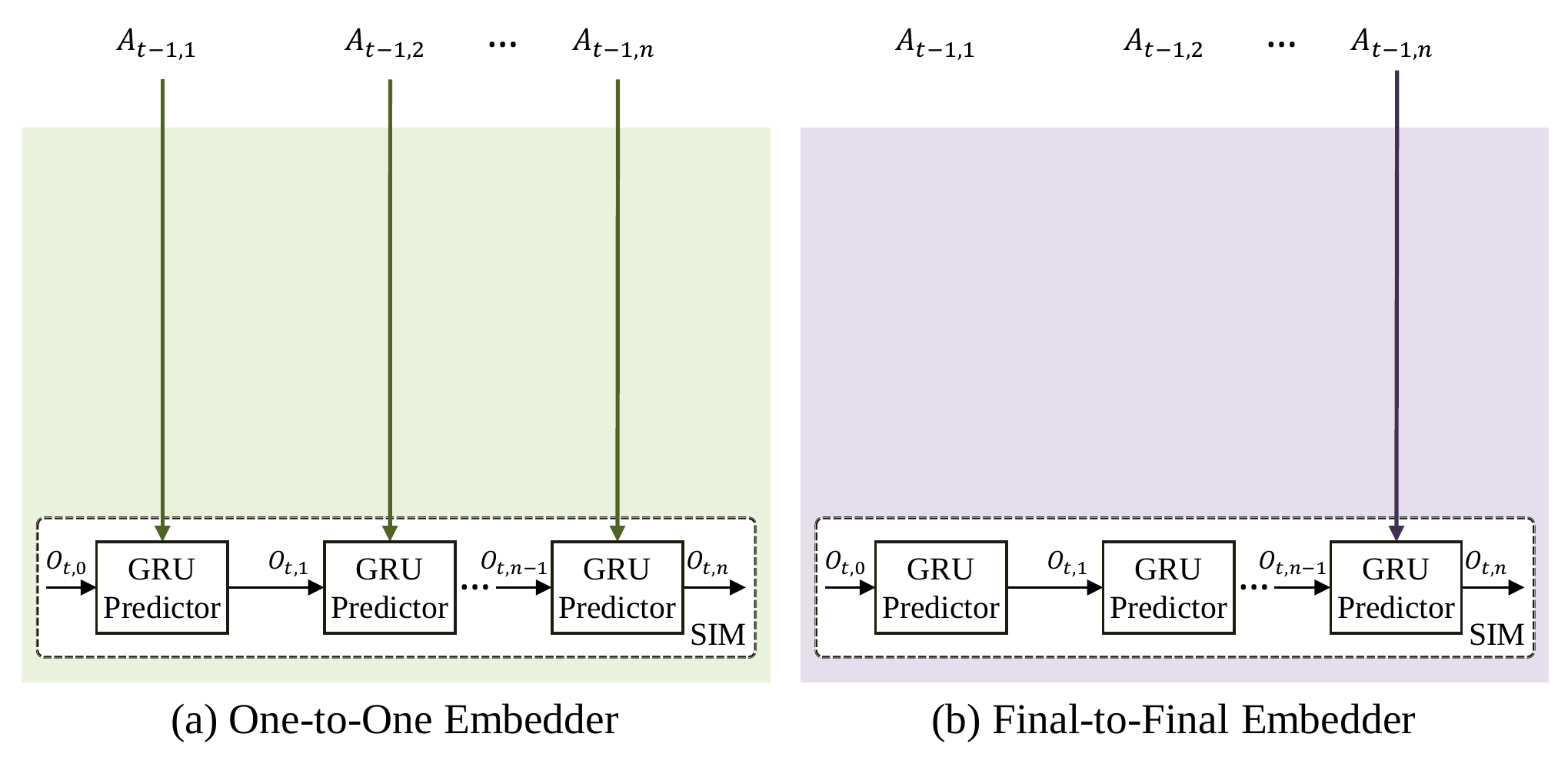}
	\end{center}
	\caption{
	(a) The schematic diagrams of the One-to-One embedding method. (b) The schematic diagrams of the Final-to-Final embedding method. $O_{t,n}$ is the frame $t-1$'s estimated low-resolution optical flow. $A_{t-1,n}$ is the $n$th iteration's aligned motion feature.
	}
	\label{fig:embedder}
\end{figure}

%% file: fig/iter.tex
\begin{figure}[h]
	\begin{center}
	    \includegraphics[width=\linewidth]{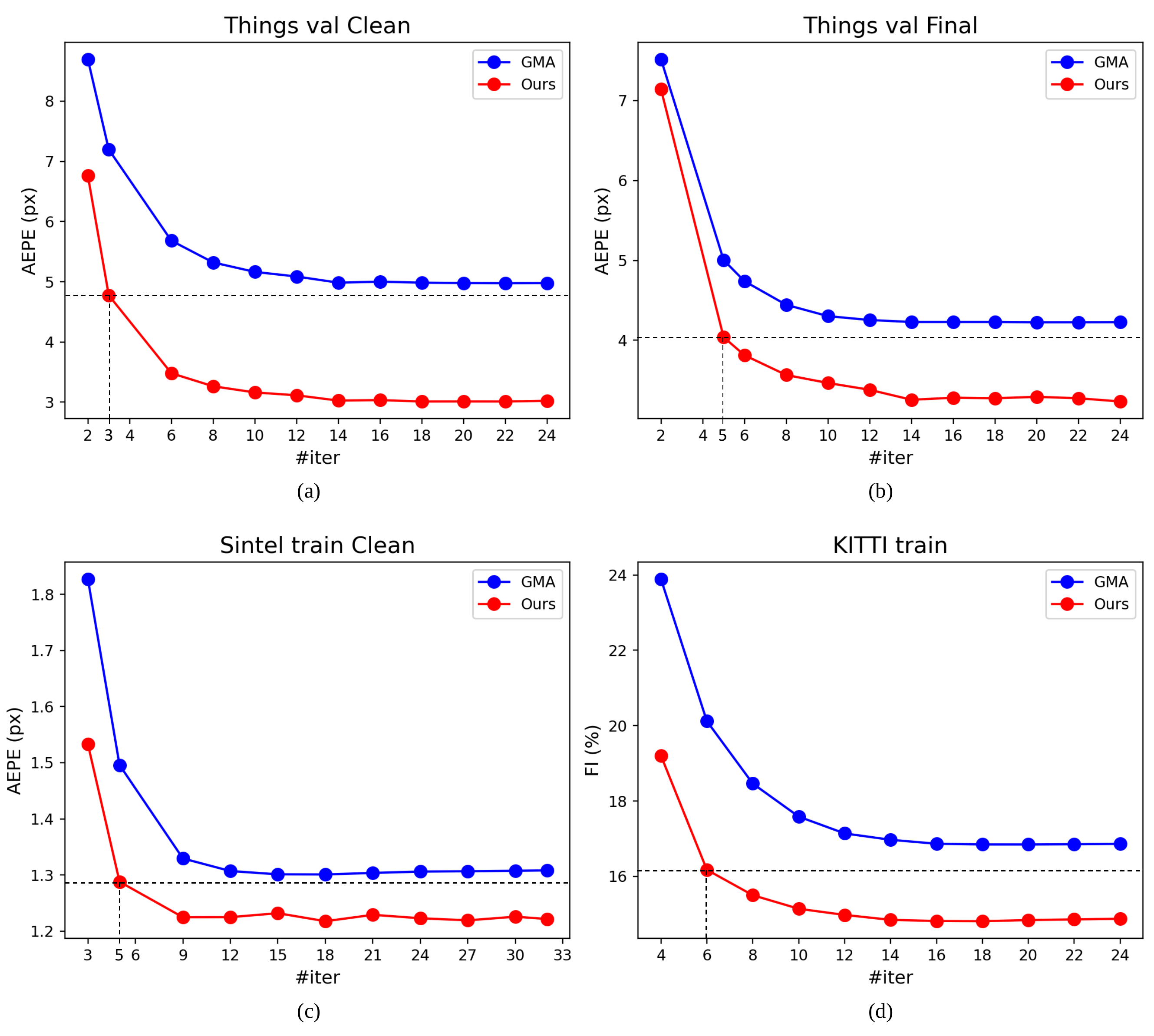}
	\end{center}
	\caption{
    Datasets evaluation results vs. the number of iterations at inference time. Our method and GMA \cite{GMA} quickly converge as single-resolution iterative methods, comparable with final results only after 12 iterations. Our method surpasses GMA \cite{GMA}'s final performance only after less than five average iterations.
	}
	\label{fig:iter}
\end{figure}

%% file: table/multi-frame.tex
\begin{table*}[t]
\centering
\caption{
Quantitative results of comparisons among multi-frame methods. All data come from the C+T training process. The data marked with parentheses under the ``Parameters'' column indicates the model parameter amount used for inferring non-first video frames.
}
\label{tab:multi-frame}
\resizebox{0.7\textwidth}{!}{%
\begin{tabular}{cccccccccc}
\toprule
\multirow{3}{*}{Backbone} & \multirow{3}{*}{\begin{tabular}[c]{@{}c@{}}Multi-frame\\ Method\end{tabular}} & \multicolumn{2}{c}{Things \cite{Things}} & \multicolumn{2}{c}{Sintel \cite{Sintel}} & \multicolumn{2}{c}{KITTI \cite{KITTI}} & \multirow{3}{*}{FPS} & \multirow{3}{*}{Parameters} \\ \cmidrule(lr){3-4} \cmidrule(lr){5-6} \cmidrule(lr){7-8}
 &  & \multicolumn{2}{c}{val} & \multicolumn{2}{c}{train} & \multicolumn{2}{c}{train} &  &  \\
 &  & Clean & Final & Clean & Final & EPE & Fl-all &  &  \\ \midrule \midrule
\multirow{4}{*}{RAFT \cite{RAFT}} & - & 5.24 & 4.74 & 1.46 & 2.70 & 5.03 & 17.5 & 17.06 & \textbf{5.3M} \\ \cmidrule{2-10}
 & FusionFlow \cite{ren2019fusion} & 4.82 & 5.11 & 1.36 & 2.97 & 4.89 & 17.7 & 4.87 & 5.8M \\
 & WarmStart \cite{RAFT} & 4.96 & 4.64 & 1.42 & \textbf{2.70} & 4.92 & 17.5 & 13.10 & \textbf{5.3M} \\
 & {\ul SplatFlow (Ours)} & \textbf{3.43} & \textbf{3.29} & \textbf{1.24} & 2.84 & \textbf{4.55} & \textbf{16.2} & \textbf{15.35} & 8.0M/(5.7M) \\ \midrule
\multirow{4}{*}{GMA \cite{GMA}} & - & 4.98 & 4.23 & 1.31 & 2.75 & 4.48 & 16.9 & 12.96 & \textbf{5.9M} \\ \cmidrule{2-10}
 & FusionFlow \cite{ren2019fusion} & 4.60 & 5.02 & 1.29 & 3.16 & 4.47 & 16.8 & 4.16 & 6.4M \\
 & WarmStart \cite{RAFT} & 5.16 & 4.24 & 1.30 & \textbf{2.71} & 4.48 & 16.9 & 10.61 & \textbf{5.9M} \\
 & {\ul SplatFlow (Ours)} & \textbf{3.02} & \textbf{3.23} & \textbf{1.22} & 2.97 & \textbf{3.70} & \textbf{14.9} & \textbf{12.29} & 9.0M/(6.3M) \\ \bottomrule
\end{tabular}
}
\end{table*}

%% file: fig/frame_num.tex
\begin{figure}[h]
	\begin{center}
		\includegraphics[width=\linewidth]{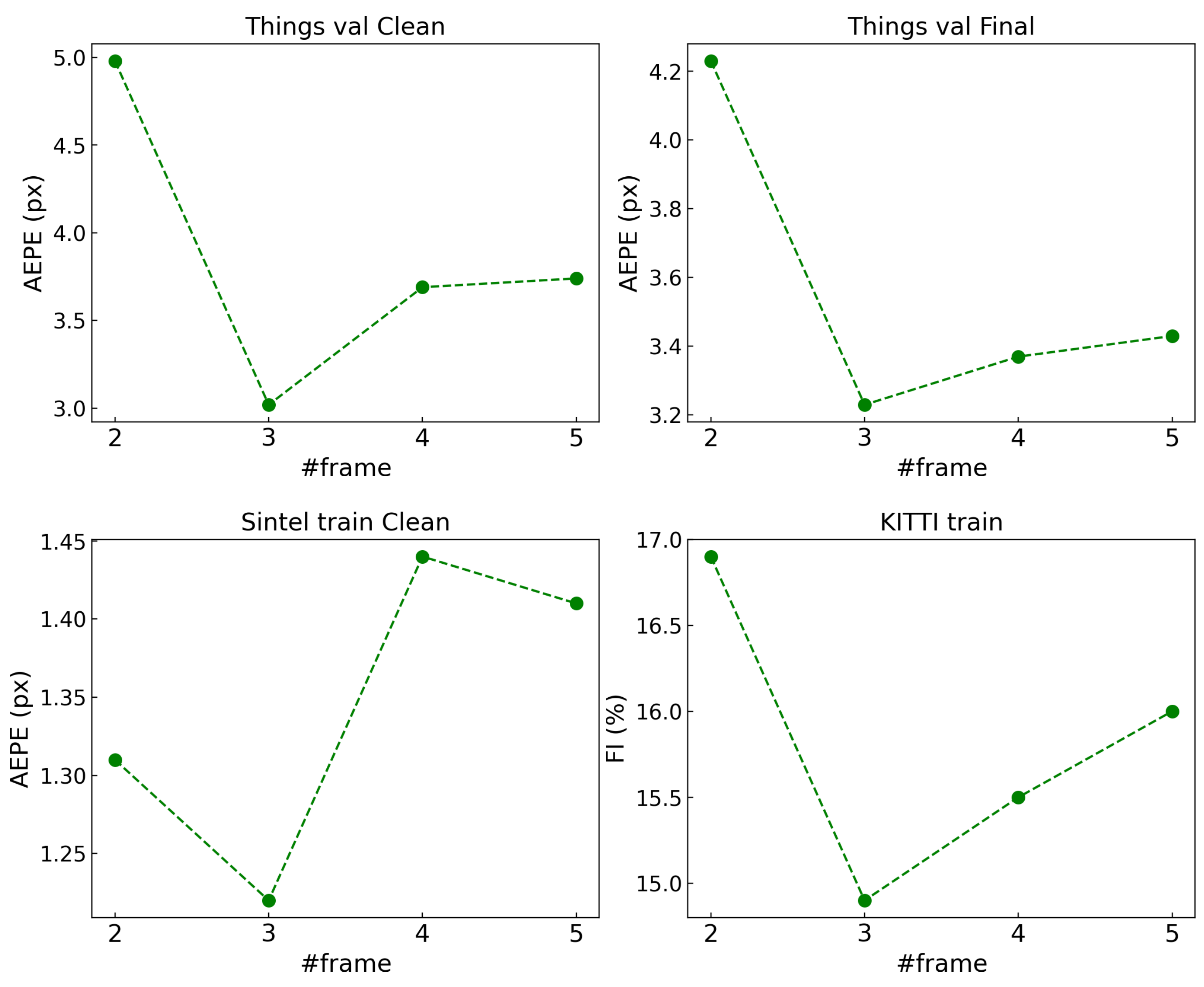}
	\end{center}
	\caption{
	Datasets evaluation results vs. the number of frames. A point with a horizontal axis of 2 represents GMA \cite{GMA}, and a point with a horizontal axis of 3/4/5 represents the 3/4/5-frame version of the proposed SplatFlow. All data come from the C+T training process. 
	}
	\label{fig:frame_num}
\end{figure}

%% file: table/multi-occ.tex
\begin{table}[h]
\centering
\caption{
The evaluation results of multi-frame methods on three types of regions (non-occluded, occluded, and all) on the Things \cite{Things} dataset. All data come from the C+T training process. 
The data marked with parentheses indicate the increments compared with the backbones (\textcolor{red}{red} indicates the error increases, and \textcolor{blue}{blue} indicates the error decreases)
}
\label{tab:multi-occ}
\begin{tabular}{lrrr}
\toprule
\multicolumn{1}{c}{\multirow{2}{*}{Method}} &
  \multicolumn{3}{c}{Things \cite{Things} val Clean} \\ \cmidrule(lr){2-4} 
\multicolumn{1}{c}{} &
  \multicolumn{1}{c}{Noc} &
  \multicolumn{1}{c}{Occ} &
  \multicolumn{1}{c}{All} \\ \midrule \midrule
RAFT \cite{RAFT} &
  1.88 &
  16.62 &
  5.24 \\
+FusionFlow \cite{ren2019fusion} &
  1.80 (\textcolor{blue}{$-$0.08}) &
  15.27 (\textcolor{blue}{$-$1.35}) &
  4.82 (\textcolor{blue}{$-$0.42}) \\
+WarmStart \cite{RAFT} &
  1.90 (\textcolor{red}{$+$0.02}) &
  15.49 (\textcolor{blue}{$-$1.13}) &
  4.96 (\textcolor{blue}{$-$0.28}) \\
{\ul +SplatFlow (Ours) } &
  \textbf{1.43 (\textcolor{blue}{$-$0.45})} &
  \textbf{8.85 (\textcolor{blue}{$-$7.77})} &
  \textbf{3.43 (\textcolor{blue}{$-$1.81})} \\ \midrule
GMA \cite{GMA} &
  1.96 &
  15.59 &
  4.98 \\
+FusionFlow \cite{ren2019fusion} &
  1.82 (\textcolor{blue}{$-$0.14}) &
  13.95 (\textcolor{blue}{$-$1.64}) &
  4.60 (\textcolor{blue}{$-$0.38}) \\
+WarmStart \cite{RAFT} &
  2.11 (\textcolor{red}{$+$0.15}) &
  15.56 (\textcolor{blue}{$-$0.03}) &
  5.16 (\textcolor{red}{$+$0.18}) \\
{\ul +SplatFlow (Ours) } &
  \textbf{1.32 (\textcolor{blue}{$-$0.64})} &
  \textbf{8.05 (\textcolor{blue}{$-$7.54})} &
  \textbf{3.02 (\textcolor{blue}{$-$1.96})} \\ \bottomrule
\end{tabular}
\end{table}

%% file: table/two-frame.tex
\begin{table*}[t]
\centering
\caption{Quantitative results of comparisons between the proposed method and two-frame backbone networks. All data come from the C+T training process. The data marked with parentheses under the ``Parameters'' column indicates the model parameter amount used for inferring non-first video frames.}
\label{tab:two-frame}
\resizebox{0.8\textwidth}{!}{%
\begin{tabular}{cccccccccc}
\toprule
\multirow{3}{*}{Backbone} & \multirow{3}{*}{\begin{tabular}[c]{@{}c@{}}Multi-frame\\ Settings\end{tabular}} & \multicolumn{2}{c}{Things \cite{Things}} & \multicolumn{2}{c}{Sintel \cite{Sintel}} & \multicolumn{2}{c}{KITTI \cite{KITTI}} & \multirow{3}{*}{FPS} & \multirow{3}{*}{Parameters} \\
 &  & \multicolumn{2}{c}{val} & \multicolumn{2}{c}{train} & \multicolumn{2}{c}{train} &  &  \\ \cmidrule(lr){3-4} \cmidrule(lr){5-6} \cmidrule(lr){7-8}
 &  & Clean & Final & Clean & Final & EPE & Fl-all &  &  \\ \midrule \midrule
\multirow{2}{*}{RAFT \cite{RAFT}} & No & 5.24 & 4.74 & 1.46 & \textbf{2.70} & 5.03 & 17.5 & \textbf{17.06} & \textbf{5.3M} \\
 & {\ul Yes (SplatFlow-RAFT)} & \textbf{3.43} & \textbf{3.29} & \textbf{1.24} & 2.84 & \textbf{4.55} & \textbf{16.2} & 15.35 & 8.0M/(5.7M) \\ \midrule
\multirow{2}{*}{GMA \cite{GMA}} & No & 4.98 & 4.23 & 1.31 & \textbf{2.75} & 4.48 & 16.9 & \textbf{12.96} & \textbf{5.9M} \\
 & {\ul Yes (SplatFlow-GMA)} & \textbf{3.02} & \textbf{3.23} & \textbf{1.22} & 2.97 & \textbf{3.70} & \textbf{14.9} & 12.29 & 9.0M/(6.3M) \\ \bottomrule
\end{tabular}
}
\end{table*}

%% file: fig/occ.tex
\begin{figure*}[t]
	\begin{center}
	    \includegraphics[width=0.9\linewidth]{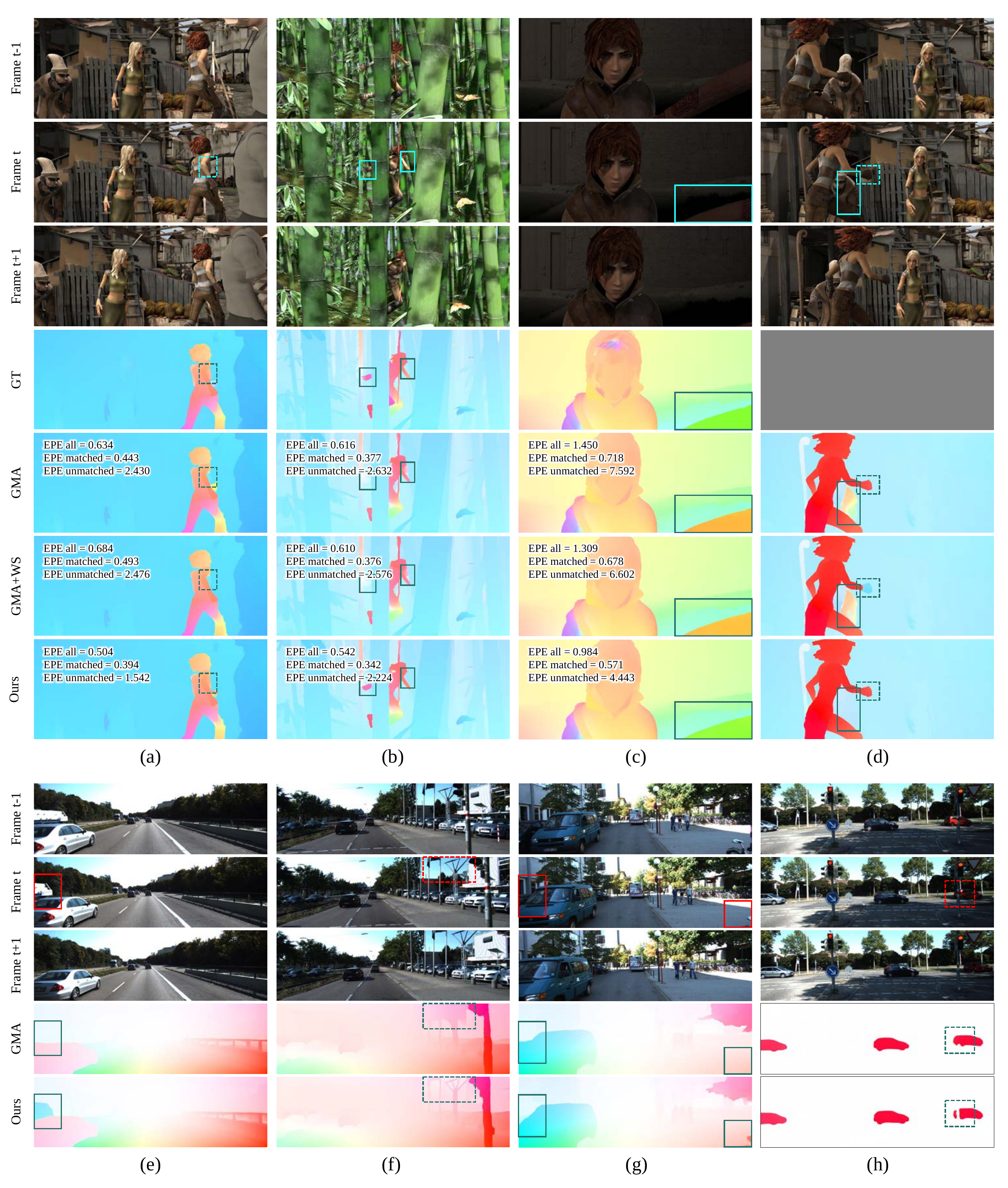}
	\end{center}
	\caption{
	Qualitative results of GMA \cite{GMA} and our method on the Sintel \cite{Sintel} test Clean ((a)-(d)) and KITTI \cite{KITTI} test ((e)-(h)) datasets.
	The ``GMA+WS'' row in (a)-(d) shows the qualitative results of the WarmStart \cite{RAFT} method on the Sintel \cite{Sintel} test Clean dataset.
	The (a)-(c) columns provide examples, including the ground truths and evaluation values reported on the Sintel benchmark. Contrarily, the ground truth of (d) is unavailable and is replaced by gray. The solid-box marked regions are significantly occluded in the frame $t+1$, and the dotted-box marked regions are non-occluded but challenging to estimate. The box contents show that our method can obtain more detailed results in non-occluded regions while achieving more acceptable performance in occluded regions.
	}
	\label{fig:occ}
\end{figure*}

%% file: table/occ.tex
\begin{table*}[t]
\centering
\caption{The evaluation results and relative improvements compared the proposed methods with their two-frame backbones on three types of regions (non-occluded, occluded, and all).}
\label{tab:occ}
\resizebox{0.9\textwidth}{!}{%
\begin{tabular}{ccccccccc}
\toprule
\begin{tabular}[c]{@{}c@{}}Training\\ Process\end{tabular} & Dataset & \begin{tabular}[c]{@{}c@{}}Region\\ Type\end{tabular} & RAFT \cite{RAFT} & \begin{tabular}[c]{@{}c@{}}SplatFlow\\ -RAFT\end{tabular} & \begin{tabular}[c]{@{}c@{}}Rel. Impr.\\ on RAFT\\ (\%)\end{tabular} & GMA \cite{GMA} & \begin{tabular}[c]{@{}c@{}}SplatFlow\\ -GMA\end{tabular} & \begin{tabular}[c]{@{}c@{}}Rel. Impr.\\ on GMA\\ (\%)\end{tabular} \\ \midrule
\multirow{6}{*}{C+T} & \multirow{3}{*}{\begin{tabular}[c]{@{}c@{}}Things \cite{Things} val\\ Clean\end{tabular}} & Noc & 1.88 & \textbf{1.43} & 23.9 & 1.96 & \textbf{1.32} & 32.7 \\
 &  & Occ & 16.62 & \textbf{8.85} & \textbf{46.8} & 15.59 & \textbf{8.05} & \textbf{48.4} \\
 &  & All & 5.24 & \textbf{3.43} & 34.5 & 4.98 & \textbf{3.02} & 39.4 \\ \cmidrule(lr){2-9}
 & \multirow{3}{*}{\begin{tabular}[c]{@{}c@{}}Things \cite{Things} val\\ Final\end{tabular}} & Noc & 1.72 & \textbf{1.45} & 15.7 & 1.71 & \textbf{1.53} & 10.5 \\
 &  & Occ & 14.90 & \textbf{8.27} & \textbf{44.5} & 13.41 & \textbf{8.24} & \textbf{38.6} \\
 &  & All & 4.74 & \textbf{3.29} & 30.6 & 4.23 & \textbf{3.23} & 23.6 \\ \cmidrule(lr){2-9}
 & \multirow{3}{*}{\begin{tabular}[c]{@{}c@{}}Sintel \cite{Sintel} train\\ Clean\end{tabular}} & Noc & 0.64 & \textbf{0.61} & 4.7 & 0.59 & \textbf{0.59} & 0.0 \\
 &  & Occ & 11.81 & \textbf{9.24} & \textbf{21.8} & 10.49 & \textbf{9.23} & \textbf{12.0} \\
 &  & All & 1.46 & \textbf{1.24} & 15.1 & 1.31 & \textbf{1.22} & 6.9 \\ \midrule
\multirow{12}{*}{Sintel-finetune} & \multirow{3}{*}{\begin{tabular}[c]{@{}c@{}}Sintel \cite{Sintel} train\\ Clean\end{tabular}} & Noc & 0.33 & \textbf{0.28} & 15.2 & 0.29 & \textbf{0.26} & 10.3 \\
 &  & Occ & 6.23 & \textbf{4.24} & \textbf{31.9} & 4.96 & \textbf{4.00} & \textbf{19.4} \\
 &  & All & 0.76 & \textbf{0.57} & 25.0 & 0.63 & \textbf{0.53} & 15.9 \\ \cmidrule(lr){2-9}
 & \multirow{3}{*}{\begin{tabular}[c]{@{}c@{}}Sintel \cite{Sintel} train\\ Final\end{tabular}} & Noc & 0.67 & \textbf{0.54} & 19.4 & 0.58 & \textbf{0.53} & 8.6 \\
 &  & Occ & 8.13 & \textbf{6.10} & \textbf{25.0} & 7.04 & \textbf{5.74} & \textbf{18.5} \\
 &  & All & 1.22 & \textbf{0.94} & 23.0 & 1.05 & \textbf{0.91} & 13.3 \\ \cmidrule(lr){2-9}
 & \multirow{3}{*}{\begin{tabular}[c]{@{}c@{}}Sintel \cite{Sintel} test\\ Clean\end{tabular}} & Noc & 0.76 & \textbf{0.51} & 32.9 & 0.56 & \textbf{0.51} & 8.9 \\
 &  & Occ & 11.48 & \textbf{6.62} & \textbf{42.3} & 8.13 & \textbf{6.06} & \textbf{25.5} \\
 &  & All & 1.93 & \textbf{1.18} & 38.9 & 1.39 & \textbf{1.12} & 19.4 \\ \cmidrule(lr){2-9}
 & \multirow{3}{*}{\begin{tabular}[c]{@{}c@{}}Sintel \cite{Sintel} test\\ Final\end{tabular}} & Noc & 1.55 & \textbf{1.38} & 11.0 & 1.40 & \textbf{1.06} & 24.3 \\
 &  & Occ & 16.46 & \textbf{12.85} & \textbf{21.9} & 14.96 & \textbf{10.29} & \textbf{31.2} \\
 &  & All & 3.18 & \textbf{2.64} & 17.0 & 2.88 & \textbf{2.07} & 28.1 \\ \bottomrule
\end{tabular}
}
\end{table*}

%% file: table/benchmarks.tex
\begin{table*}[!h]
\centering
\caption{Quantitative results on Sintel \cite{Sintel} and KITTI \cite{KITTI} benchmarks. The best result of each metric is marked in bold. The evaluation result on the training dataset of the training process is marked in parentheses, and the evaluation result on the testing dataset is not marked in parentheses. * indicates the result with the WarmStart \cite{RAFT} initialization method of single-resolution iterative networks.}
\label{tab:benchmarks}
\resizebox{0.85\textwidth}{!}{%
\begin{tabular}{ccccccccc}
\toprule
\multirow{2}{*}{Training Data} & \multirow{2}{*}{Method} & \multicolumn{2}{c}{Sintel \cite{Sintel} train} & \multicolumn{2}{c}{KITTI \cite{KITTI} train} & \multicolumn{2}{c}{Sintel \cite{Sintel} test} & KITTI \cite{KITTI} test\\ \cmidrule(lr){3-4} \cmidrule(lr){5-6} \cmidrule(lr){7-8} \cmidrule(lr){9-9}
 &  & Clean & Final & AEPE & F1-all & Clean & Final & Fl-all \\ \midrule \midrule
\multirow{9}{*}{C+T} & LiteFlowNet2 \cite{hui2020lightweight} & 2.24 & 3.78 & 8.97 & 25.9 & - & - & - \\
 & VCN \cite{yang2019volumetric} & 2.21 & 3.68 & 8.36 & 25.1 & - & - & - \\
 & MaskFlowNet \cite{zhao2020maskflownet} & 2.25 & 3.61 & - & 23.1 & - & - & - \\
 & FlowNet2 \cite{Flownet2} & 2.02 & 3.54 & 10.08 & 30 & 3.96 & 6.02 & - \\
 & DICL \cite{wang2020displacement} & 1.94 & 3.77 & 8.7 & 23.6 & - & - & - \\
 & RAFT \cite{RAFT} & 1.43 & 2.71 & 5.04 & 17.4 & - & - & - \\
 & SeparableFlow \cite{zhang2021separable} & 1.30 & \textbf{2.59} & 4.60 & 15.9 & - & - & - \\
 & GMA \cite{GMA} & 1.30 & 2.74 & 4.69 & 17.1 & - & - & - \\
 & Ours & \textbf{1.22} & 2.97 & \textbf{3.70} & \textbf{14.9} & - & - & - \\ \midrule
\multirow{13}{*}{S-finetune} & FlowNet2 \cite{Flownet2} & (1.45) & (2.01) & - & - & 4.16 & 5.74 & - \\
 & PWC-Net+ \cite{sun2019models} & (1.71) & (2.34) & - & - & 3.45 & 4.60 & - \\
 & LiteFlowNet2 \cite{hui2020lightweight} & (1.30) & (1.62) & - & - & 3.48 & 4.69 & - \\
 & HD3 \cite{yin2019hierarchical} & (1.87) & (1.17) & - & - & 4.79 & 4.67 & - \\
 & IRR-PWC \cite{hur2019iterative} & (1.92) & (2.51) & - & - & 3.84 & 4.58 & - \\
 & VCN \cite{yang2019volumetric} & (1.66) & (2.24) & - & - & 2.81 & 4.40 & - \\
 & MaskFlowNet \cite{zhao2020maskflownet} & - & - & - & - & 2.52 & 4.17 & - \\
 & ScopeFlow \cite{Scopeflow} & - & - & - & - & 3.59 & 4.10 & - \\
 & DICL \cite{wang2020displacement} & (1.11) & (1.60) & - & - & 2.12 & 3.44 & - \\
 & RAFT \cite{RAFT} & (0.76) & (1.22) & - & - & 1.61* & 2.86* & - \\
 & SeparableFlow \cite{zhang2021separable} & (0.69) & (1.10) & - & - & 1.50 & 2.67 & - \\
 & GMA \cite{GMA} & (0.62) & (1.06) & - & - & 1.39* & 2.47* & - \\
 & Ours & \textbf{(0.53)} & \textbf{(0.91)} & - & - & \textbf{1.12} & \textbf{2.07} & - \\ \midrule
\multirow{13}{*}{K-finetune} & FlowNet2 \cite{Flownet2} & - & - & (2.30) & (6.8) & - & - & 11.48 \\
 & PWC-Net+ \cite{sun2019models} & - & - & (1.50) & (5.3) & - & - & 7.72 \\
 & LiteFlowNet2 \cite{hui2020lightweight} & - & - & (1.47) & (4.8) & - & - & 7.74 \\
 & HD3 \cite{yin2019hierarchical} & - & - & (1.31) & (4.1) & - & - & 6.55 \\
 & IRR-PWC \cite{hur2019iterative} & - & - & (1.63) & (5.3) & - & - & 7.65 \\
 & VCN \cite{yang2019volumetric} & - & - & (1.16) & (4.1) & - & - & 6.30 \\
 & MaskFlowNet \cite{zhao2020maskflownet} & - & - & - & - & - & - & 6.10 \\
 & ScopeFlow \cite{Scopeflow} & - & - & - & - & - & - & 6.82 \\
 & DICL \cite{wang2020displacement} & - & - & (1.02) & (3.6) & - & - & 6.31 \\
 & RAFT \cite{RAFT} & - & - & (0.63) & (1.5) & - & - & 5.10 \\
 & SeparableFlow \cite{zhang2021separable} & - & - & (0.69) & (1.6) & - & - & 4.64 \\
 & GMA \cite{GMA} & - & - & \textbf{(0.57)} & \textbf{(1.2)} & - & - & 5.15 \\
 & Ours & - & - & (0.80) & (2.4) & - & - & \textbf{4.61} \\ \bottomrule
\end{tabular}%
}
\end{table*}

%% file: table/final.tex
\begin{table*}[t]
\centering
\caption{Quantitative results of comparisons between the proposed method and its backbone GMA \cite{GMA} after the C+T training process on the Sintel \cite{Sintel} train Final dataset sequence by sequence. The ``Error Increment'' row only displays an error increment whose value is a positive number. The largest error increment is marked in bold.
}
\label{tab:final}
\resizebox{0.8\linewidth}{!}{%
\begin{tabular}{ccccccccc}
\toprule \midrule
Scene Name & alley\_1 & alley\_2 & ambush\_2 & ambush\_4 & ambush\_5 & ambush\_6 & ambush\_7 & bamboo\_1 \\ \midrule
GMA & \textbf{0.21} & 0.20 & \textbf{20.44} & \textbf{14.63} & \textbf{6.98} & \textbf{7.42} & 0.52 & 0.38 \\
{\ul Ours} & 0.21 & \textbf{0.17} & 20.70 & 21.94 & 8.20 & 8.12 & \textbf{0.38} & \textbf{0.36} \\
Error Increment & 0.00 & - & 0.26 & \textbf{7.31} & 1.22 & 0.70 & - & - \\ \midrule \midrule
Scene Name & bamboo\_2 & bandage\_1 & bandage\_2 & cave\_2 & cave\_4 & market\_2 & market\_5 & market\_6 \\ \midrule
GMA & 0.78 & 0.42 & \textbf{0.46} & 5.81 & 3.15 & \textbf{0.60} & 9.03 & 2.34 \\
{\ul Ours} & \textbf{0.66} & \textbf{0.40} & 0.46 & \textbf{5.10} & \textbf{3.15} & 0.63 & \textbf{8.88} & \textbf{1.87} \\
Error Increment & - & - & 0.00 & - & - & 0.03 & - & - \\ \midrule \midrule
Scene Name & mountain\_1 & shaman\_2 & shaman\_3 & sleeping\_1 & sleeping\_2 & temple\_2 & temple\_3 &  \\ \midrule
GMA & \textbf{0.26} & 0.26 & 0.24 & 0.12 & 0.11 & 2.33 & \textbf{4.01} &  \\
{\ul Ours} & 0.28 & \textbf{0.25} & \textbf{0.22} & \textbf{0.11} & \textbf{0.09} & \textbf{1.84} & 4.34 &  \\
Error Increment & 0.02 & - & - & - & - & - & 0.33 &  \\ \bottomrule
\end{tabular}
}
\end{table*}

%% file: fig/final_error.tex
\begin{figure}[h]
	\begin{center}
		\includegraphics[width=\linewidth]{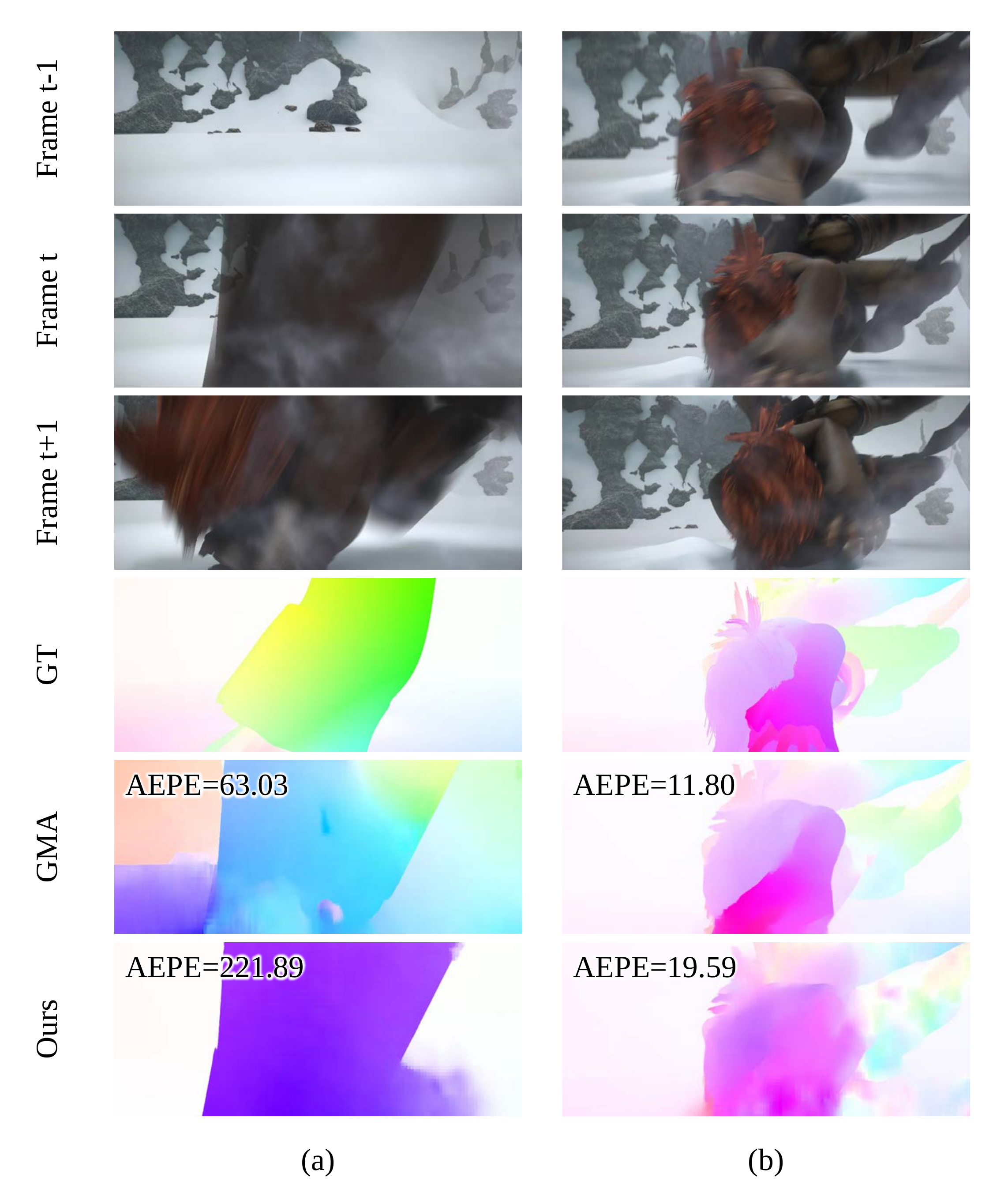}
	\end{center}
	\caption{
	Qualitative results of two cases in which our method has large errors on the ``ambush\_4'' sequence of the Sintel \cite{Sintel} train Final dataset. In the first case (a), the image is seriously blurred, and the motion displacement is enormous; it cannot be judged even by human eyes. In the other case (b), the image has large blurs, which are not found during training.
	}
	\label{fig:final_error}
\end{figure}

%% file: body/tail.tex
\section*{Acknowledgement}
This work was partially supported by National Key Research and Development Program of China No. 2021YFB3100800, the National Natural Science Foundation of China under Grant 61973311, 62376283 and 62006239, the Defense industrial Technology Development Program (JCKY2020550B003) and the Key Stone grant (JS2023-03) of the National University of Defense Technology (NUDT).

\paragraph{\textbf{Conflict of interest}} The authors declare that they have no conflict of interest.

\paragraph{\textbf{Data Availability}} All datasets used are publicly available. Code is available at \href{https://github.com/wwsource/SplatFlow}{https://github.com/wwsource/SplatFlow}.

\bibliographystyle{spbasic}      
\footnotesize
\bibliography{wangbo}